%% file: main.tex
\documentclass[10pt,twocolumn,letterpaper]{article}

\usepackage{cvpr}              
\usepackage[accsupp]{axessibility}

\input{preamble}
\definecolor{cvprblue}{rgb}{0.21,0.49,0.74}
\usepackage[pagebackref,breaklinks,colorlinks,allcolors=cvprblue]{hyperref}
\usepackage{algorithm}
\usepackage{algorithmic}

\def\paperID{3816} 
\def\confName{CVPR}
\def\confYear{2026} 

\title{GardenDesigner: Encoding Aesthetic Principles into Jiangnan Garden Construction via a Chain of Agents} 
\author{Mengtian Li\textsuperscript{1,2}, Fan Yang\textsuperscript{1}, Ruixue Xiong\textsuperscript{1}, Yiyan Fan\textsuperscript{1}, Zhifeng Xie\textsuperscript{1,2$\dagger$}, Zeyu Wang\textsuperscript{3$\dagger$}\\
\textsuperscript{1}Shanghai University\\
\textsuperscript{2}Shanghai Engineering Research Center of Motion Picture Special Effects\\
\textsuperscript{3}The Hong Kong University of Science and Technology (Guangzhou)\\
{\tt\small \{mtli, yangphan, xiongruixue, yiyanfan, zhifeng\_xie\}@shu.edu.cn, zeyuwang@ust.hk}
}

\begin{document}

\twocolumn[{%
\renewcommand\twocolumn[1][]{#1}%
\maketitle

\includegraphics[width=\linewidth]{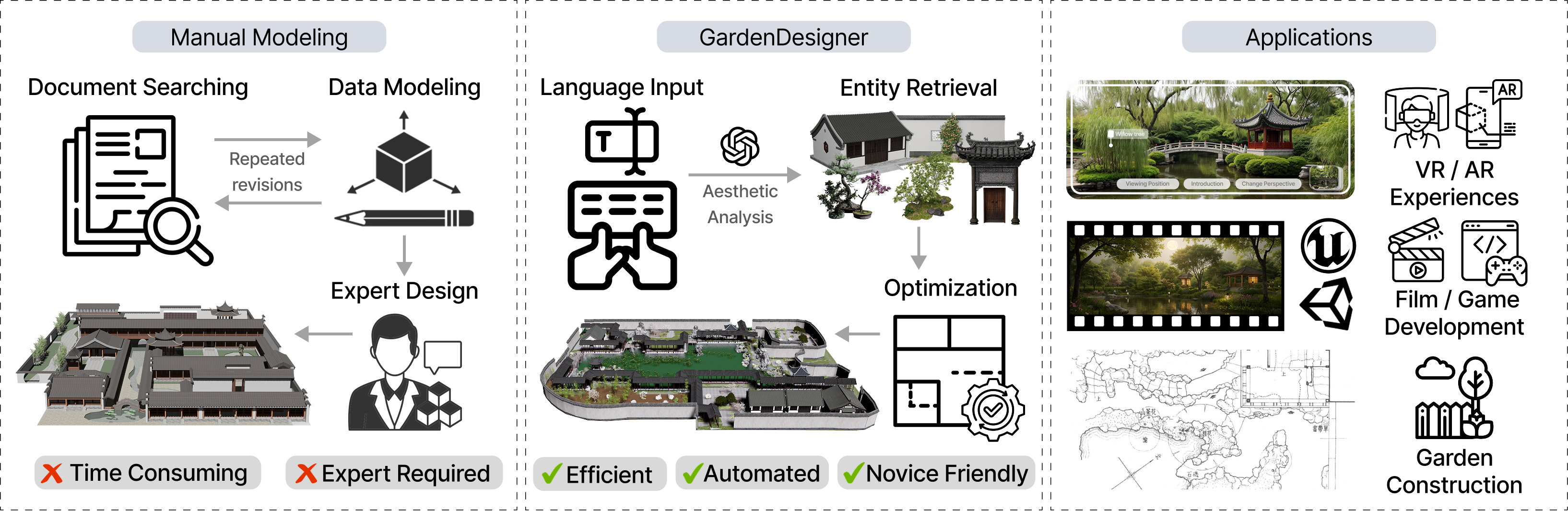}
\captionof{figure}{
The motivation of GardenDesigner. Traditional manual modeling of Jiangnan gardens requires document searching, data modeling, and expert design, making it time-consuming and expertise-dependent. GardenDesigner automates Jiangnan garden construction
via analyzing the user text and acquiring the assets, and then optimizes the garden layout. For applications, users can construct a Jiangnan garden through text input, which can be used for creating VR/AR experiences, film and game development, and real garden construction.\vspace{1em}}
\label{fig:teaser}
}]

\renewcommand\thefootnote{}
\footnotetext{$^\dagger$ Corresponding authors.}

\input{Sec/0_abstract}
\input{Sec/1_introduction}
\input{Sec/2_related_work}

\input{Sec/3_method}

\input{Sec/4_dataset}

\input{Sec/5_experiment}
\input{Sec/6_conclusion}

\section{Acknowledgments} 
This work was supported by the National Natural Science Foundation of China (Grant No. 62402306), the Natural Science Foundation of Shanghai (Grant No. 24ZR1422400, Grant No. 25ZR1401130), the Open Research Project of the State Key Laboratory of Industrial Control Technology, China (Grant No. ICT2024B72), and the Guangdong Basic and Applied Basic Research Foundation (No. 2026A1515011138).

{
    \small
    \bibliographystyle{ieeenat_fullname}
    \bibliography{main}
}


\input{Sec/7_supp}

\end{document}

%% file: Sec/0_abstract.tex
\begin{abstract}
Jiangnan gardens, a prominent style of Chinese classical gardens, hold great potential as digital assets for film and game production and digital tourism.
However, manual modeling of Jiangnan gardens heavily relies on expert experience for layout design and asset creation, making the process time-consuming.
To address this gap, we propose \textbf{GardenDesigner}, a novel framework that encodes aesthetic principles for Jiangnan garden construction and integrates a chain of agents based on procedural modeling.
The water-centric terrain and explorative pathway rules are applied by terrain distribution and road generation agents. Selection and spatial layout of garden assets follow the aesthetic and cultural constraints. Consequently, we propose asset selection and layout optimization agents to select and arrange objects for each area in the garden.
Additionally, we introduce \textbf{GardenVerse} for Jiangnan garden construction, including expert-annotated garden knowledge to enhance the asset arrangement process.
To enable interaction and editing, we develop an interactive interface and tools in Unity, in which non-expert users can construct Jiangnan gardens via text input within one minute.
Experiments and human evaluations demonstrate that GardenDesigner can generate diverse and aesthetically pleasing Jiangnan gardens. Project page is available at \url{https://monad-cube.github.io/GardenDesigner}.
\end{abstract}

%% file: Sec/1_introduction.tex
\section{Introduction}
As the most important genres of Chinese classical gardens, Jiangnan gardens exemplify compact urban compositions with intricate spatial configurations~\cite{yuanye}. Unlike general landscape parks, they emphasize a balance of architecture, plants, and rocks. Typical features include winding corridors, attics, pavilions that frame ever-changing views, rockeries that simulate mountains within limited space, and ponds that reflect both natural scenery and surrounding structures~\cite{peng1986analysis}. 
The traditional construction of Jiangnan gardens involves much manual effort, including three main steps: (1) document search, collecting historical documents, drawings, and photographs; (2) asset modeling, reconstructing architectural elements and plants based on these materials; (3) expert design, addressing terrain shaping and garden layout, relying on specialized knowledge. 
However, this process typically involves three to four designers and takes about three to four weeks to complete, making it heavily reliant on manual effort and time-consuming.

Current learning-based scene generation methods~\cite{text2room,commonscenes,meng2024lt3sd} exhibit limited generalizability due to domain constraints in training datasets.
Procedural modeling methods~\cite{holodeck,sun2025layoutvlm,conlan} that incorporate large language models (LLMs) or visual language models (VLMs) focus on either spatially limited room space or unstructured natural environments.
However, the construction of Jiangnan gardens remains unexplored, and three problems remain to be addressed.
\textbf{(1) Complex terrain and garden layout}: Compared to general landscapes, Jiangnan gardens exhibit intricate terrain structures and spatial layouts, where terrain, water, and architecture are interwoven under implicit aesthetic logic.
\textbf{(2) Aesthetic principle constraints}: Due to the abstract nature of Jiangnan gardens' design rules, encoding the aesthetic principles into a computational generation framework remains challenging.
\textbf{(3) Absence of Jiangnan garden dataset}: Lacking stylistic appearance and cultural annotation, existing 3D datasets of ordinary or urban objects are not suitable for Jiangnan garden construction.

To address these challenges, we propose \textbf{GardenDesigner}, which integrates a chain of agents, procedural modeling, and aesthetic principles encoding for Jiangnan gardens construction. 
Specifically, GardenDesigner is composed of three modules: Hierarchical Garden Composition (Section~\ref{method: Hierarchical Garden Composition}), Knowledge-embedded Asset Arrangement (Section~\ref{method: Knowledge-Embedded Asset Arrangement}). 
First, Hierarchical Garden Composition decomposes the construction process into procedural terrain and road generation.
Subsequently, Knowledge-embedded Asset Arrangement focuses on asset selection and optimizing objects according to the specified constraints for each area. 
The key insight is to select objects and set constraints according to area information and expert-guided garden knowledge. Consequently, we introduce \textbf{GardenVerse}, a high-quality Jiangnan garden dataset that contains typical Jiangnan garden style of digital assets with expert-annotated garden knowledge, enhancing the specific knowledge context for knowledge-embedded asset arrangement.

To support convenient designing and interaction, we develop an interface and editing tools in Unity, in which the non-expert user can construct Jiangnan gardens via text input within one minute. After construction, the system can output the 2D garden layout as a reference for the real garden creation and building. In summary, GardenDesigner opens new avenues for intangible cultural heritage preservation and creative applications in digital art and games.

Our main contributions are as follows: 
\begin{itemize}
  \item We propose \textbf{GardenDesigner}, a novel framework that encodes aesthetic principles for Jiangnan garden construction via integrating a chain of agents with an expert-annotated artistic dataset \textbf{GardenVerse}.
  \item We propose a hierarchical garden composition module to generate terrain and roads with aesthetic principles, and a knowledge-embedded asset arrangement mechanism for asset selection and layout optimization.
  \item We develop an interface and editing tools in Unity, in which non-expert users can construct Jiangnan gardens via text input. The system outputs a 2D layout for real garden construction and supports virtual tourism.
\end{itemize} 

 

%% file: Sec/2_related_work.tex
\section{Related Work}

\subsection{Scene Generation} 

\noindent\textbf{Procedural Scene Generation.} Procedurally generating scenes with rules and manual algorithms has long been a robust methodology. CityEngine~\cite{parish2001procedural} and~\citet{khan2019procsy} procedurally model the city. 
Recently,~\citet{infinigen, infinigenindoors}  generates assets in scenes from shape to texture. 

\noindent\textbf{Data-Driven Scene Generation.} 
Previous learning-based methods have explored different modalities to generate scenes, including images~\cite{wonderworld}, texts~\cite{text2room}, layouts~\cite{bahmani2023cc3d}, scene graphs~\cite{commonscenes} and raw room~\cite{CharacterRoom}, while
some methods~\cite{xie2024citydreamer,xie2024gaussiancity} extend
to large-scale city generation. 

\noindent\textbf{Scene Generation with LLMs and VLMs.} 
~\citet{layoutgpt} takes the first step to utilize LLMs to generate object position, while some methods~\cite{anyhome, holodeck,i-design} generate the scene graph.
Other methods~\cite{3d-gpt,scenecraft,zhou2024scenex,liu2025worldcraft} explore the outdoor generation based on Blender~\cite{blender} or Infinigen~\cite{infinigen}. ~\citet{conlan} procedurally model the landscape with LLMs.
~\citet{layoutgpt} adapt VLM to optimize indoor layout and some methods~\cite{holodeck2.0, ling2025scenethesis} explored simple outdoor scene.                         

Previous methods have primarily focused on ordinary indoor spaces or unstructured landscapes. In contrast, generating Jiangnan gardens poses unique challenges, requiring fine-grained spatial composition, hierarchical reasoning, and the integration of aesthetic and cultural principles. 


\subsection{3D Object Datasets}

\noindent\textbf{Indoor and Ordinary Objects.} ShapeNet~\cite{shapenet} collects 3D CAD models from public repositories and previous datasets. 
GSO~\cite{gso} offers scans of household objects, and OmniObject3D~\cite{omniobject3d} expands both quantity and diversity. 
Objaverse-XL~\cite{objaverse-xl} extends Objaverse~\cite{objaverse} to 10.2M 3D assets. 
However, existing datasets lack sufficient diversity or fidelity for cultural scenes such as Jiangnan gardens.

\noindent\textbf{Outdoor and Natural Objects.} 
BuildingNet~\cite{buildingnet} and CityCraft~\cite{deng2024citycraft} mine the architectures from websites~\cite{sketchfab,3dwarehouse}. 
~\citet{zhu2024crops3d} collects a scanned 3D crops dataset and some methods ~\cite{zhou2024scenex,le2021eden} create architectural or natural assets with Unreal~\cite{unreal} or Blender~\cite{blender}.
Procedural modeling methods~\cite{parish2001procedural,khan2019procsy} employ parametric or L-system rules for virtual cities.
Other works~\cite{li2024interactive,wong2016procedural} simulate vegetation, and Infinigen~\cite{infinigen} extends to large-scale natural textured assets.

Despite extensive research on architectural and natural objects, Jiangnan gardens featuring traditional architecture, distinctive flora, and rocks remain underexplored. Existing datasets lack the stylistic coherence, cultural context, and fine-grained diversity needed for heritage-oriented scenes.

\subsection{Cultural Heritage and Digital Tourism}
Cultural Heritage (CH) encompasses tangible and intangible artifacts, traditions, and environments, in which interactive systems transform preservation from passive documentation to active participation.
To foster public engagement, prior works have explored diverse cultural heritage applications, including immersive cultural tourism~\cite{kim2009AR}, underwater heritage exploration~\cite{zhang2020deep}, and interactive historical storytelling and artifact preservation~\cite{jamil2019augmented}.

These applications effectively employ immersion, narrative, and interaction to represent CH. However, most focus on heritage exploration and exhibition rather than the generation of heritage-inspired content.
%
Therefore, a generative and interactive system is essential to lower the creative threshold, translating complex garden aesthetics into tangible designs through simple text input.
Such an approach not only preserves the Jiangnan garden tradition but also revitalizes it as a living, participatory form of cultural heritage.


%% file: Sec/3_method.tex
\begin{figure*}[th]
  \centering
  \includegraphics[width=\textwidth]{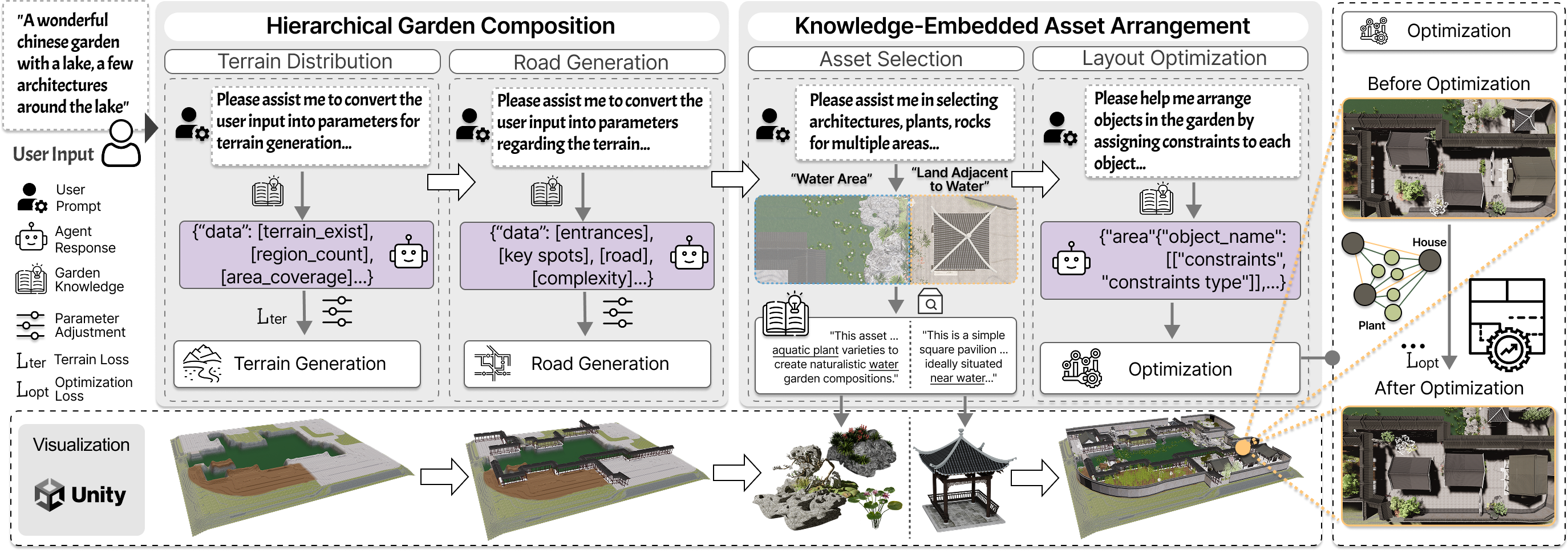}
  \caption{Overview of the GardenDesigner pipeline. 
  GardenDesigner transforms the user input into a Jiangnan garden through Hierarchical Garden Composition and Knowledge-Embedded Asset Arrangement. 
  First, Hierarchical Garden Composition transfers the user input into parameters for terrain and road generation with aesthetic principles. Subsequently, Knowledge-Embedded Asset Arrangement chooses the objects based on the garden knowledge and area information, 
  and then optimization loss is used to get the feasible solution for layout.
  }
  \label{fig:GardenDesigner}
\end{figure*}

\section{Method}
\subsection{Problem Statement}
Jiangnan garden construction involves generating terrain and roads, configuring objects in a bounded space based on the user's instructions, and following certain Jiangnan garden design rules.
After integrating expert experience of garden designers and literature search~\cite{yuanye, peng1986analysis}, we summarize key aesthetic principles to guide Jiangnan garden construction from four perspectives of \textit{terrain distribution}, \textit{road generation}, \textit{asset selection}, and \textit{relational constraint}:
\begin{itemize}
    \item \textbf{Naturalistic and Water-Centric Foundation:} The terrain is designed to be an idealized, miniature microcosm of a natural landscape, and the water is considered the lifeblood and the soul of the garden to organize elements. 

    \item \textbf{Discovery and Winding Paths:} Following the water and area border, paths are designed for exploration, creating a series of unfolding, painterly scenes rather than simple transit, deliberately avoiding straight lines and symmetry.

    \item \textbf{Symbolism and Miniature:} Selects assets that are symbolic miniatures of the natural world. The assets should be culturally appropriate and fit their specific location, reflecting both natural content and cultural intentionality.

    \item \textbf{Asymmetrical Balance:} Arranges objects harmoniously, creating a dynamic and natural balance. Positional constraints are used to hide and reveal views, making the garden feel larger and encouraging exploration.
\end{itemize}

Formally, given a user text $U$, aesthetic principles $K_\text{global}$ and garden assets $O_\text{asset}=\{o_{1},...,o_{n}\}$ with knowledge annotation $K_{a}=\{k_{1},...,k_{n}\}$, the objective is to create a Jiangnan garden that meets the textual input and aesthetic principles. 
We decompose the Jiangnan garden construction task into four steps and implement it via a chain of agents, in which the content generated by previous agents serves as the basis for subsequent agents.
(1) \textbf{Terrain Distribution Agent ($\mathcal{A}_T$):} this agent generates the terrain $T$ based on the user's text and global knowledge.
(2) \textbf{Road Generation Agent ($\mathcal{A}_R$):} based on terrain $T$, this agent generates the road $R$, also guided by global knowledge. 
(3) \textbf{Asset Selection Agent ($\mathcal{A}_S$):} with terrain and road context $(T, R)$ available, this agent selects a set of appropriate assets $O_{s}$ from the library. 
(4) \textbf{Layout Optimization Agent ($\mathcal{A}_C$):} the final agent takes the terrain, paths, and selected objects as input, arranges the selected objects, optimizing their position and rotation.
Therefore, the complete garden $G$ is the composite output derived from the chain of agents:
\begin{equation}
G = (T, R, (O_\text{s}, P)),
\end{equation}
where selected objects $O_{s}=\{o_{1},...,o_{m}\}$ with properties $P = ((x_{i}, y_{i}, z_{i}, r_{i}),...)$, representing position and rotation.

\subsection{Hierarchical Garden Composition}
\label{method: Hierarchical Garden Composition}
\noindent\textbf{Challenges.}
(1) \textit{Water-centric spatial organization.}
Conventional landscape procedural algorithms fail to capture the water-centered logic of Jiangnan gardens.
As a result, they often produce scattered ponds and unnatural terrain that disrupt the intended harmony between land and water.
(2) \textit{Exploratory path generation.}
Existing path-generation methods focus on geometric efficiency or uniform coverage, neglecting the exploratory routing principles of Jiangnan gardens.
Thus, they cannot reproduce the winding, layered paths that define the authentic visitor experience.

To address these challenges, we introduce two agents: (1) \textbf{Terrain Distribution Agent}  $A_\text{T}$ and (2) \textbf{Road Generation Agent} $A_\text{R}$. 
These agents leverage specific garden composition prompts, integrate a water-centric loss to guide the generation and optimization of terrain, and redesign the road scoring mechanism to encourage roads to follow terrain boundaries and avoid excessive linearity.

\textbf{Genetic Terrain Generation.}
The Jiangnan garden is generally located in flat terrain areas within urban areas and occupies relatively small sites.
Consequently, we adopt a genetic algorithm based on a 2D grid and choose four types of terrain to simulate the landform of Jiangnan garden: \textit{Outside}, \textit{Waterbody}, \textit{Land}, and \textit{Ground}, represented as integer numbers. 
To enable language control, $A_{T}$ is used to generate terrain, which transfers the text input to the parameters and then calls the genetic algorithm with the parameters:
\begin{equation}
T = \mathcal{A}_\text{T}(U, K_\text{global}),
\end{equation}
where $U$ is the user input text, and $K_\text{global}$ is the aesthetic principles.
Specifically,  we choose four types of terrain parameters: (1) existence, (2) quantity, (3) coverage, and (4)  single region coverage. 
Based on the parameters, the genetic algorithm conducts the \textit{Crossover}, \textit{Mutation}, and \textit{Evolution} operations for each iteration. Finally, the fitness function is used to select the feasible terrain solution for each iteration.
Most importantly, we introduce a water-centric loss to calculate the terrain fitness as follows:
\begin{equation}
L_{\textnormal{terrain}}=f_{}\cdot\max(1-\frac{\sum_{i=0}^{n} c(T,(x_{i},y_{i}))}{\phi},0),
\end{equation}
where $T$ represents the generated terrain, $f$ is the factor, $c$ function is used to judge whether the grid is in water.

\textbf{Explorative Road Generation.}
Given the discretized terrain layout, $A_{R}$ synthesizes roads adhering to Jiangnan aesthetic principles. 
We integrate cultural priors into a 
grid-based scoring mechanism and produces smooth spline curves for a realistic pedestrian experience and corridor arrangement.
First, the agent parses the user instruction $U$ to generate the parameters, including the number of entrances and keypoints, the width of the main road, and the road complexity, which jointly determine the roads and entrances of the garden. The entrances are sampled across all directional boundaries, and then the roads are generated by scoring the grid border and selecting the best solution. Additionally, the path selection process follows the Jiangnan garden key requirements: (1) the roads can reach most of the garden area, (2) the roads prefer to follow the border, and (3) the roads should avoid excessive warping and straightening. The process is formulated as follows: 
\begin{equation}
R = \mathcal{A}_\text{R}(\mathcal{S}(T,e_{i,j}), U, K_\text{global}),
\end{equation}
where $e$ is the edge in the grid and $S$ is the scoring function according to the rules and principles.

\subsection{Knowledge-Embedded Asset Arrangement}
\label{method: Knowledge-Embedded Asset Arrangement}
\noindent\textbf{Challenges.}
(1) \textit{Rule-based and aesthetic spatial logic.}
Conventional retrieval or constraint methods fail to capture implicit culturally grounded relations in Jiangnan gardens, leading to aesthetically inconsistent layouts.
(2) \textit{Lack of domain-specific understanding.}
General LLMs lack garden knowledge, making it hard to reason about the interplay of architectural, structural, and botanical elements, thus failing to produce layouts aligned with traditional design logic.

To tackle these challenges, we first annotate the garden asset dataset with descriptions encoding expert garden knowledge. Then, we propose a knowledge-embedded asset arrangement mechanism, consisting of knowledge-guided asset retrieval and aesthetic constraints encoding, implemented by the \textbf{Asset Selection Agent} $A_\text{S}$ and the \textbf{Layout Optimization Agent} $A_\text{C}$.

\subsubsection{Knowledge-Guided Asset Retrieval}  
First, we collected a Jiangnan garden dataset, GardenVerse, and then proposed a knowledge-guided agent $A_\text{S}$ to retrieve assets with expert-annotated garden knowledge.
Specifically, we annotated the object assets with additional garden knowledge description $K_{a} = \{k_{1},...,k_{n}\}$ to provide the agents with rich knowledge about the garden objects in Section~\ref{sec: dataset}. We encode these annotations into a knowledge vector store and query them through a large language model to enforce culturally consistent object selection. To get appropriate objects, we provide the area information $I_{\textnormal{area}} = \{i_{1},...i_{k}\}$ and garden knowledge for LLM, then agent will response a list of object $O_{\textnormal{s}}=\{o_{1},...o_{m}\}$ for object arrangement of each area, as follows:
\begin{equation}
O_\text{s} = \mathcal{A}_\text{S}(\mathcal{Q}((\mathcal{V}(K_\text{a}), o_{i}), U), I_\text{area}),
\end{equation}
where $\mathcal{Q}$ is the query operation, $\mathcal{V}$ is the process to vector store, $o_{i}$ refers to each object and $i \in \{0,...m\}$.

\subsubsection{Aesthetic Constraints Encoding} 
\label{Aesthetic Constraints Encoding.}
To address the challenge of inconsistent aesthetic constraints, we set the constraints for selected objects and then optimize the layout according to the constraints. 
Specifically, we define eight constraint types and group them into five semantic categories according to their spatial position, direction relationship with the boundary and objects: (1) Global (edge, middle) indicates the overall placement within the entire scene; (2) Position (around, backed up) captures relative placement relationships; (3) Distance (near, far) quantifies spatial proximity; (4) Alignment (aligned) enforces consistent directional orientation among objects; and (5) Rotation (face to) specifies the facing direction of an object toward another.

\begin{figure}[th]
  \centering
\includegraphics[width=0.47\textwidth]{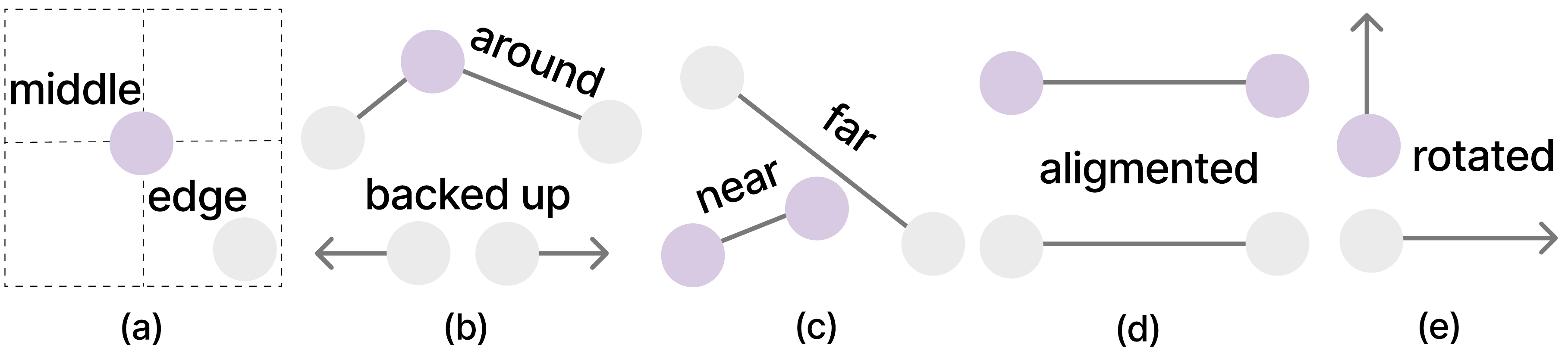}
  \caption{The five constraints categories: (a) Global, edge, and middle;
  (b) Position, around, and backed up;
  (c) Distance, near and far;
  (d) Alignment, aligned;
  And (e) Rotation, face to.
  }
  \label{fig:method_constraint}
\end{figure}

\textbf{Optimization.}
To generate the garden layout, we design five types of optimization loss functions, corresponding to different categories of spatial constraints. 
The position and direction for each object is represented as $o_{i}=(x_{i},y_{i},z_{i},\theta_{i})$ and the bounding box is $b_{i}=(l_{i},w_{i},h_{i})$. We formulate the optimization loss as follows.

\textit{Global Objective} is used to decide the global position and optimize objects to the edge or middle of an area:
{\small
\begin{equation}
\mathcal{L}_{\text{glo}} = 
\begin{cases}
\max\left(\frac{d( o_{i}, e_\text{area}) - d_\text{e}}{d_\text{e}},0\right), & \text{if edge}, \\
\max\left(\frac{\left\| o_{i} - c_\text{area} \right\| - d_\text{m}}{d_\text{m}}, 0\right), & \text{if middle},
\end{cases}
\end{equation}
}
where $e_\text{area}$ and $c_\text{area}$ are the boundary and the center, $d_\text{e}$ and $d_\text{m}$ are the threshold value parameters. The $d$ is used to calculate the distance between a point and an area boundary. 
  
\textit{Position Objective} loss focuses on the relative position and direction between two different objects:
{\small
\begin{equation}
\mathcal{L}_{\text{pos}} = 
\begin{cases}
m\left(r_\text{l} - d, 0\right) + m\left(d - r_\text{h}, 0\right), & \text{if around}, \\
f_{\text{back}} \cdot f(o_{i}, o_{j}, \theta), & \text{if backed up},
\end{cases}
\end{equation}
}where $m$ is the max function, $r_\text{l}$ and $r_\text{h}$ represent the low and high threshold. $d$ is the distance between two objects. $f_{\text{back}}$ is the parameter, $\theta$ is the front orientation of $o_{j}$ and $f$ is deciding if $o_{i}$ is backed up $o_{j}$.

\textit{Distance Objective} is used to control and adjust the relative distance between different objects:
{\small
\begin{equation}
\mathcal{L}_{\text{dis}} = 
\begin{cases}
\max\left(\frac{\| o_{i} - o_{j} \|-d_\text{n}}{d_\text{n}}, 0\right), & \text{if near}, \\
\max\left(\frac{d_\text{f} - \| o_{i} - o_{j} \|}{d_\text{f}}, 0\right), & \text{if far},
\end{cases}
\end{equation}
}where $d_\text{n}$ and $d_\text{f}$ are the near and far parameters from object $o_{i}$ and another object $o_{j}$. $\epsilon$ is the threshold value parameter.

\textit{Alignment Objective} attempts to align objects of the same type for neat and regular local arrangement:
{\small
\begin{equation}
\mathcal{L}_{\text{ali}} = 
\max(\frac{|x_{i} - x_{j}|-\epsilon}{\epsilon}, 0) +\max(\frac{|y_{i} - y_{j}|-\epsilon}{\epsilon}, 0) , 
\end{equation}
}where $x$ and $y$ are the positions of two objects and $\epsilon$ represents the threshold value for alignment.

\textit{Rotation Objective} is used to adjust object direction:
{\small
\begin{equation}
\mathcal{L}_{\textnormal{rot}} = f_{\text{rot}} \cdot I(v_{i}, p(o_{j},b_{j})),
\end{equation}
}where $v_{i}$ represents the direction of $o_{i}$ and $p$ is the polygon of bounding box and position of $o_{j}$. $I$ judges whether a line and a polygon intersect. $f_{\textnormal{rot}}$ is the scale factor parameter.

And the final optimization loss is as follows:
{\small
\begin{equation}
\label{eq: final loss}
\mathcal{L}_{\text{opt}} = \lambda_1 \mathcal{L}_{\text{glo}} + \lambda_2 \mathcal{L}_{\text{pos}} + \lambda_3 \mathcal{L}_{\text{dis}} + \lambda_4 \mathcal{L}_{\text{ali}} + \lambda_5 \mathcal{L}_{\text{rot}} ,
\end{equation}
}where $\lambda_{i \in \{1,..,5 \}}$ are the loss balancing weight. The algorithm first identifies the main object and then explores placements for the anchor object. Subsequently, it employs Depth-First Search to find valid placements for the remaining objects according to the optimization loss. The whole layout optimization agent is formulated as follows: 
\begin{equation}
P = \mathcal{A}_\text{C}(\mathcal{Q}((\mathcal{V}(K_\text{a}), o_{i}, o_{j}), U)),
\end{equation}
where $o_{i}$ and $o_{j}$ are two different objects in the same area from selected objects $O_{\textnormal{s}}$, $i!=j$ and $i,j \in \{0,...m\}$.

%% file: Sec/4_dataset.tex
\section{The GardenVerse Dataset}
\label{sec: dataset}
GardenVerse comprises 132 high-quality artistic 3D assets across three canonical categories: Rock (33), Plant (44), and Architecture (54), including both individual elements ($40.2\%$) and pre-composed arrangements ($59.8\%$) of plants and rocks, enabling flexible retrieval of Jiangnan gardens.

\begin{figure}[th]
  \centering
\includegraphics[width=0.49\textwidth]{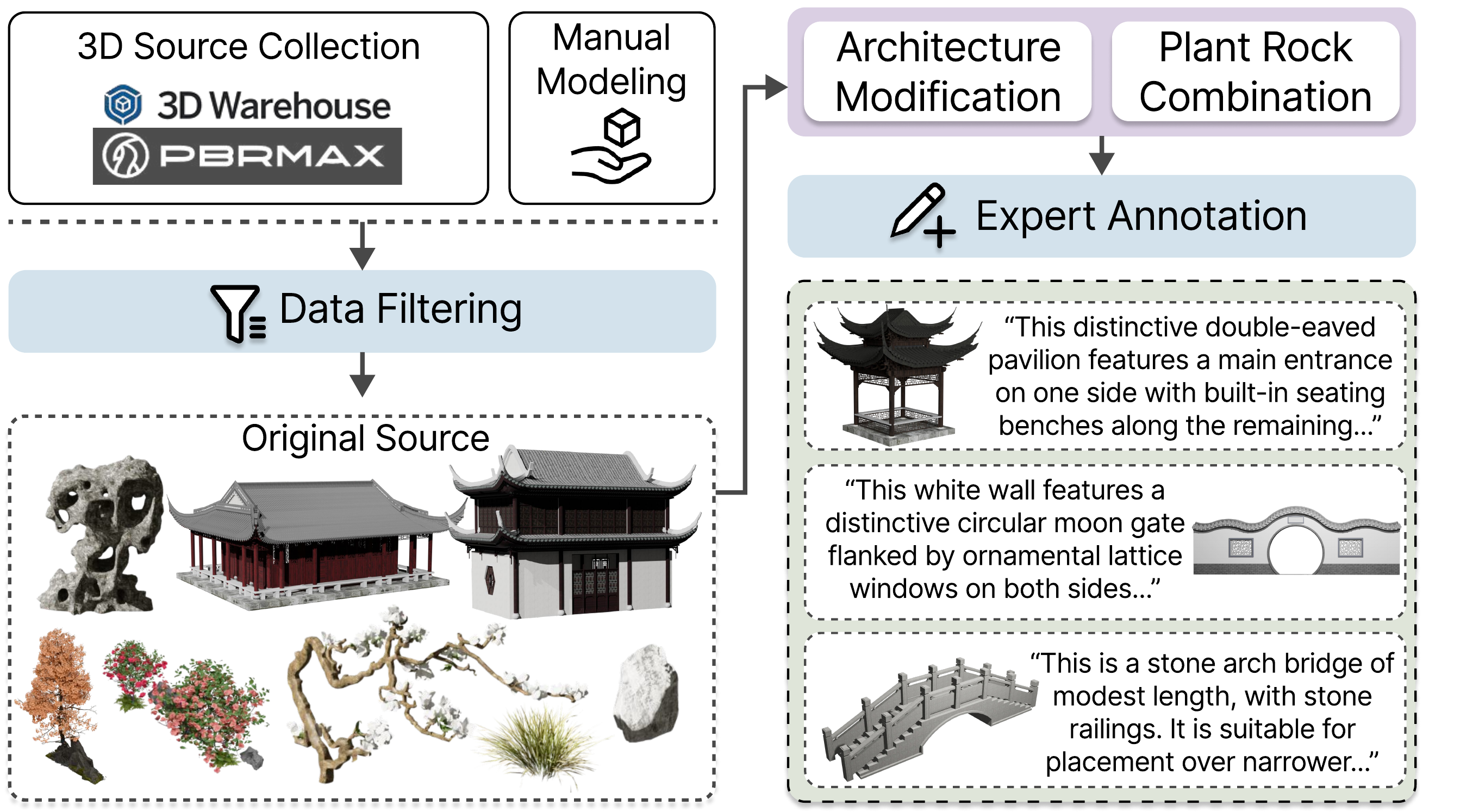}
  \caption{GardenVerse construction from Internet repositories and manual modeling. We invite experts to modify the architectures and construct object combinations. Finally, garden experts annotate the basic information and garden knowledge for assets.}
  \label{fig:data_collection}
\end{figure}

\textbf{Collection.}
We decompose four digital Jiangnan gardens into objects, including Liuyuan Garden~\cite{liuyuan}, Yiyuan Garden~\cite{yiyuan}, Wangshiyuan Garden~\cite{wangshiyuan}, and Heyuan Garden~\cite{heyuan}.
We also collect ancient architectures, plants, and rocks from the 3D Warehouse~\cite{3dwarehouse} and PBRMAX~\cite{pbrmax}. 
Then, we filter the objects with Northern garden characteristics and retain objects conforming to Jiangnan garden aesthetics. Additionally, we enforce stylistic consistency of assets through mesh optimization and material reassignment.
Also, we invite professional garden designers to create a combination of plants and rocks.

\textbf{Annotation.} After obtaining the assets, we first annotate them with basic information, including object name, size, minimum and maximum position, and related file path. Recognizing the limitations of LLMs in domain-specific tasks, we engaged landscape architecture experts to comprehensively annotate assets. Each object in GardenVerse includes detailed annotations on: visual attributes of objects, spatial compositions and arrangements, suitable season, description, and contextually appropriate placements. 
More details can be found in the supplementary materials.

%% file: Sec/5_experiment.tex
\section{Experiments}

\begin{figure*}[th]
  \centering
  \includegraphics[width=0.95\textwidth]{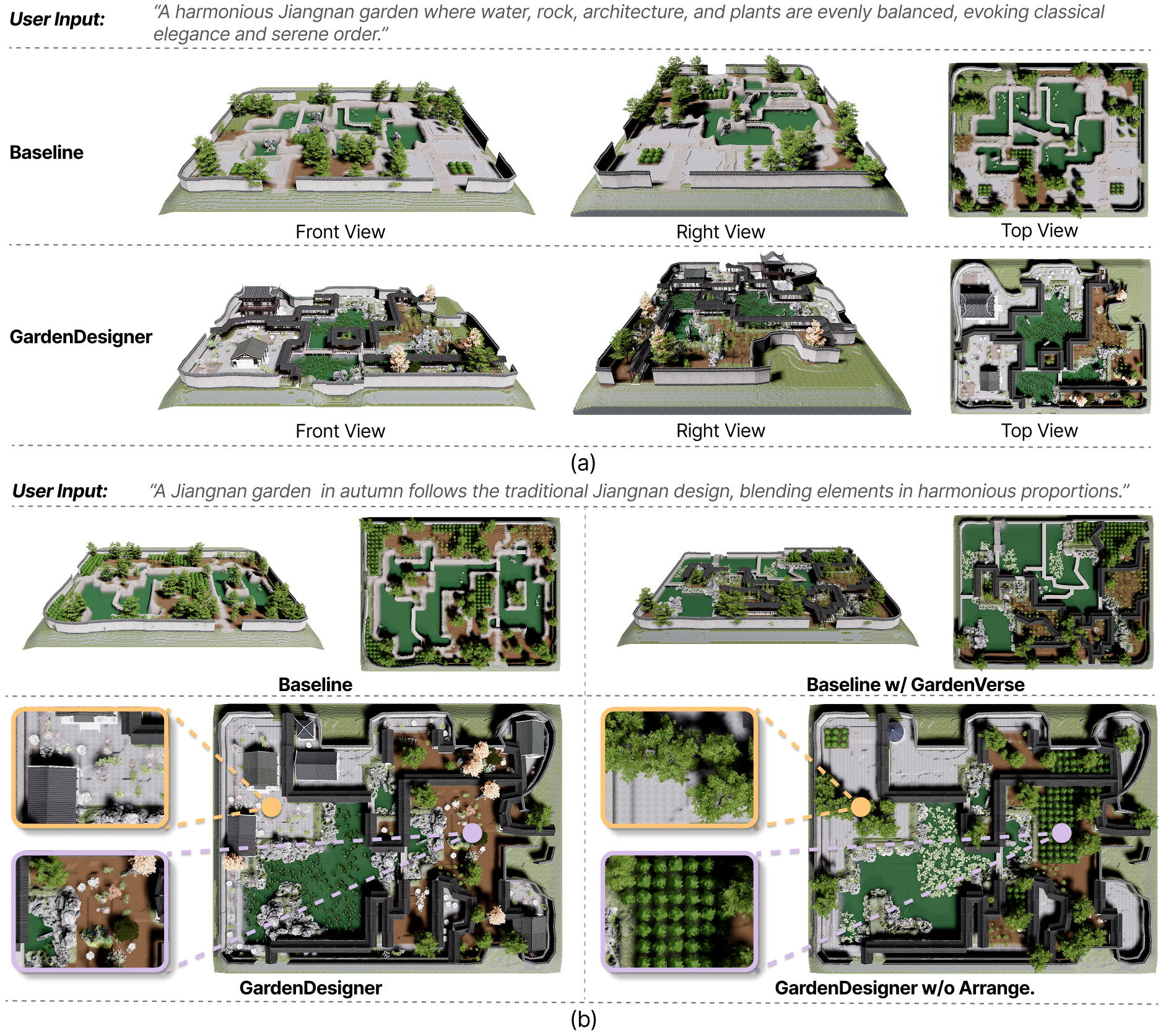}
  \caption{Qualitative analysis. In (a), we input the same prompt to GardenDesigner and the baseline~\cite{conlan} to evaluate the generated garden quality with three different views for each garden.
 In (b), we compare four methods: (1)  Baseline~\cite{conlan}, (2) Baseline with GardenVerse assets, (3) GardenDesigner, and (4) GardenDesigner without Knowledge-Embedded Asset Arrangement to conduct the ablation experiment. }
  \label{fig: Qualitative analysis}
\end{figure*}

\subsection{Experiment Setup}
\textbf{Configuration.} The garden grid is defined as $20 \times 15$ and the real garden size is defined as $200 \times 150  $ $m^{2}$. For the parameters, the weights in optimization loss are $\lambda_{i \in \{1,...,5 \}}=\{2.0, 0.5, 1.8, 0.5, 0.5\}$, and other parameter details can be found in the supplementary materials. We chose OpenAI GPT-5~\cite{gpt5} as the LLM model, file search~\cite{filesearch} for knowledge embedding, and Unity to visualize.
All reported results were obtained with an Intel(R) Core i7-13620H, 16GB memory, and NVIDIA GeForce RTX 4060 Laptop GPU.

\begin{table}[th]
  \caption{Quantitative comparison. We evaluate our method with the baseline method from four metrics: (1) the pathway rationality (Path-S), (2) the diversity of objects (Class-Div), (3) the structural complexity (FD), and (4) text and scene similarity (CLIP-S).}
  \centering
  \footnotesize
  \begin{tabular*}{0.475\textwidth}{@{\extracolsep{\fill}}llllll@{}}
    \toprule
     Method & \multicolumn{1}{c}{ Path-S $\uparrow$} & \multicolumn{1}{c}{Class-Div} & \multicolumn{1}{c}{FD}      & \multicolumn{1}{c}{CLIP-S$\uparrow$} 
     \\  \midrule
     \citet{conlan} & \multicolumn{1}{c}{0} & \multicolumn{1}{c}{21.8 $\pm$ 1.6} & \multicolumn{1}{c}{1.42 $\pm$ 0.1} & \multicolumn{1}{c}{27.4 $\pm$ 0.1}\\
    Ours 
    & \multicolumn{1}{c}{\textbf{8.1 $\pm$ 2.5}} 
    & \multicolumn{1}{c}{\textbf{68.3 $\pm$ 5.6}} 
    & \multicolumn{1}{c}{\textbf{1.36 $\pm$ 0.1}} 
    & \multicolumn{1}{c}{\textbf{27.6 $\pm$ 0.1}}\\

  \bottomrule
  \end{tabular*}
  \label{tab: quantitative comparison}
\end{table}

\begin{table}[th]
    \caption{VLMs-based comparison. We render garden images and utilize VLMs to evaluate them from rationality, aesthetic quality, and atmosphere via CLIP-A, VLM-S, and QA-Quality.}
    \footnotesize
  \centering
  \begin{tabular*}{0.475\textwidth}{@{\extracolsep{\fill}}llll@{}}
    \toprule
    Method
     & \multicolumn{1}{c}{CLIP-A $\uparrow$}
     & \multicolumn{1}{c}{VLM-S
     $\uparrow$} 
     & \multicolumn{1}{c}{QA-Quality $\uparrow$}
     \\  \midrule
     \citet{conlan}  
     & \multicolumn{1}{c}{52.9 $\pm$ 1.0}
     & \multicolumn{1}{c}{24.9 $\pm$ 1.2} 
     & \multicolumn{1}{c}{43.8 $\pm$ 2.5}
     \\
    Ours  
    & \multicolumn{1}{c}{\textbf{54.2 $\pm$ 2.0}}
    & \multicolumn{1}{c}{\textbf{32.5 $\pm$ 2.3}}
    & \multicolumn{1}{c}{\textbf{53.8 $\pm$ 3.1}}
    \\
  \bottomrule
  \end{tabular*}
    \label{tab:vlm_eval}
\end{table}

\begin{table}[th]
  \centering
    \caption{Ablation study for object layout optimization. We evaluate the Knowledge-Embedded Asset Arrangement module by removing it, based on three metric perspetives.
    }
  \footnotesize
  \begin{tabular*}{0.475\textwidth}{@{\extracolsep{\fill}}llll@{}}
    \toprule
    Method & \multicolumn{1}{c}{FD} 
     & \multicolumn{1}{c}{CLIP-S $\uparrow$} & \multicolumn{1}{c}{VLM-S $\uparrow$} \\  \midrule
    Ours w/o Arrange.  & \multicolumn{1}{c}{\textbf{1.27 $\pm$ 0.1}} & \multicolumn{1}{c}{27.4 $\pm$ 0.1} & \multicolumn{1}{c}{31.6 $\pm$ 1.1}\\
    Ours  & \multicolumn{1}{c}{1.36 $\pm$ 0.1} & \multicolumn{1}{c}{\textbf{27.6 $\pm$ 0.1}} & \multicolumn{1}{c}{\textbf{32.5 $\pm$ 2.3}}\\

  \bottomrule
  \end{tabular*}

    \label{tab:ablation}
\end{table}

\textbf{Metrics.} We evaluate generated Gardens from physical plausibility, structure complexity, semantic coherence, and aesthetic quality. 
\textit{1. Pathway Score (Path-S)}. Path-S is used to determine whether significant plants and buildings can be reached or viewed along the road:
{\small
\begin{equation}
    S_{p} = 
    \sum_{i=0}^{n} \min(\frac{d_{i}}{N}-\phi), 
\end{equation}
}where $d_{i}$ is the distance between each key spot architecture and each edge, and the $\phi$ is the threshold.
\textit{2. Class Diversity (Class-Div)}. Additionally, we also use Class Diversity to measure the object categories' diversity: 
{\small
\begin{equation}
    D_{c} = \frac{\sum_{i=0}^{n} c_{i}}{N},
\end{equation}
}where $c_{i}$ is the class number of the generated garden and $N$ is the whole asset number.
\textit{3. Fractal Dimension (FD).}   We calculate structure complexity~\cite{sun2024interpret} as follows:
{\small
\begin{equation}
D_{f}=-\lim_{r \to 0}\frac{\ln{N_{r}}}{\ln{r}},
\label{1} 
\end{equation}
}where $N_{r}$ is the number of self-similar pieces needed to cover the set at scale $r$. \textit{4. CLIP-Score} is used to measure the consistency between the generated garden and the instruction. 
\textit{5. CLIP-Aesthetic} is used to evaluate the aesthetic score. \textit{6. VLM-Score.} 
We prompt the VLMs to rate the rendered garden image. \textit{7. QA-Quality.} We also used VLM-based visual scorer Q-Align~\cite{q-align} to evaluate results.



\subsection{Qualitative Analysis}
\textbf{Comparison with the Baseline Method.} In Figure~\ref{fig: Qualitative analysis}(a), we input the same prompts for the baseline method and GardenDesigner to generate different Jiangnan gardens. The views from left to right are front, right, and top. The garden generated from the baseline method has large areas of vacancies, regular plant distribution, and no necessary architecture. In contrast, GardenDesigner gets water-centric terrain distribution, explorative road covering most garden area, and the natural garden layout. 
%
%

\textbf{Ablation Study.} 
In Figure~\ref{fig: Qualitative analysis}(b), we conduct the ablation study and select two representative areas. Compared to baseline~\cite{conlan}, GardenVerse enhances the visual quality of the other three methods, containing abundant objects and natural configurations. After removing the Knowledge-Embedded Asset Arrangement, GardenDesigner achieves a regular layout and limited objects, demonstrating the effectiveness of aesthetic principles integration. 

\subsection{Quantitative Analysis}
We compare the performance of GardenDesigner with the baseline~\cite{conlan}, as summarized in Table~\ref{tab: quantitative comparison}.
GardenDesigner achieves a higher Path-S of 8.1, reflecting more coherent relationships between architectures and roads, whereas~\citet{conlan} produces unreasonable layouts with no valid score.
In terms of asset diversity, GardenDesigner generates a wider range of garden object classes from 26 to 71 types, demonstrating greater dynamism.
For structural complexity, GardenDesigner attains a Fractal-dim of 1.36—closer to real Jiangnan gardens from 1.123 to 1.329~\cite{sun2024interpret}, indicating a more natural spatial structure.
In addition, GardenDesigner achieves a slightly higher CLIP-S of 27.6. 
Finally, we prompt VLMs with the rendered garden images and ask to rate the gardens.
In Table~\ref{tab:vlm_eval}, GardenDesigner greatly exceeds baseline~\cite{conlan} with all three aesthetic metrics.
 
\textbf{Ablation Study.} Removing the Knowledge-embedded Asset Arrangement, GardenDesigner achieves a lower garden structure complexity with 1.27, caused by fewer architectures. 
In addition, GardenDesigner gets more visual coherence 
with a CLIP-S of 27.6, 
, achieved a higher VLM-S of 32.5, validating the aesthetic quality and effectiveness. More experiments are included in supplementary materials.
%
\begin{table}[th]
  \centering
    \caption{Selection ratio (↑) of different methods for five garden types: (1) Baseline~\cite{conlan}, (2) Baseline* (Baseline with GardenVerse), (3) Ours* (GardenDesigner without Knowledge-Embedded Asset Arrangement, (4) Ours (GardenDesigner).}
  \footnotesize
  \begin{tabular*}{0.475\textwidth}{@{\extracolsep{\fill}}llllll@{}}
    \toprule
     & \multicolumn{1}{c}{Normal} & \multicolumn{1}{c}{ Hydric} & \multicolumn{1}{c}{Floral} 
     & \multicolumn{1}{c}{Arch-dense} & \multicolumn{1}{c}{Mazy} 
     \\  \midrule
    Baseline & \multicolumn{1}{c}{7.58} & \multicolumn{1}{c}{7.58} & \multicolumn{1}{c}{4.54} & \multicolumn{1}{c}{3.03} & \multicolumn{1}{c}{7.58}\\
    Baseline* & \multicolumn{1}{c}{12.12} & \multicolumn{1}{c}{9.09} & \multicolumn{1}{c}{18.18} & 
    \multicolumn{1}{c}{10.61} & 
    \multicolumn{1}{c}{22.72}\\
    Ours* & \multicolumn{1}{c}{18.18} & \multicolumn{1}{c}{40.91} & 
    \multicolumn{1}{c}{10.61} & 
    \multicolumn{1}{c}{12.12} & 
    \multicolumn{1}{c}{19.70}\\
    
    \midrule
    Ours & \multicolumn{1}{c}{62.12} & \multicolumn{1}{c}{42.42} & 
    \multicolumn{1}{c}{66.67} & 
    \multicolumn{1}{c}{74.24} & 
    \multicolumn{1}{c}{50.00}\\

  \bottomrule
  \end{tabular*}

  \label{tab:supp ablation in user study}
\end{table}

\begin{figure}[th]
  \centering
  \includegraphics[width=0.47\textwidth]{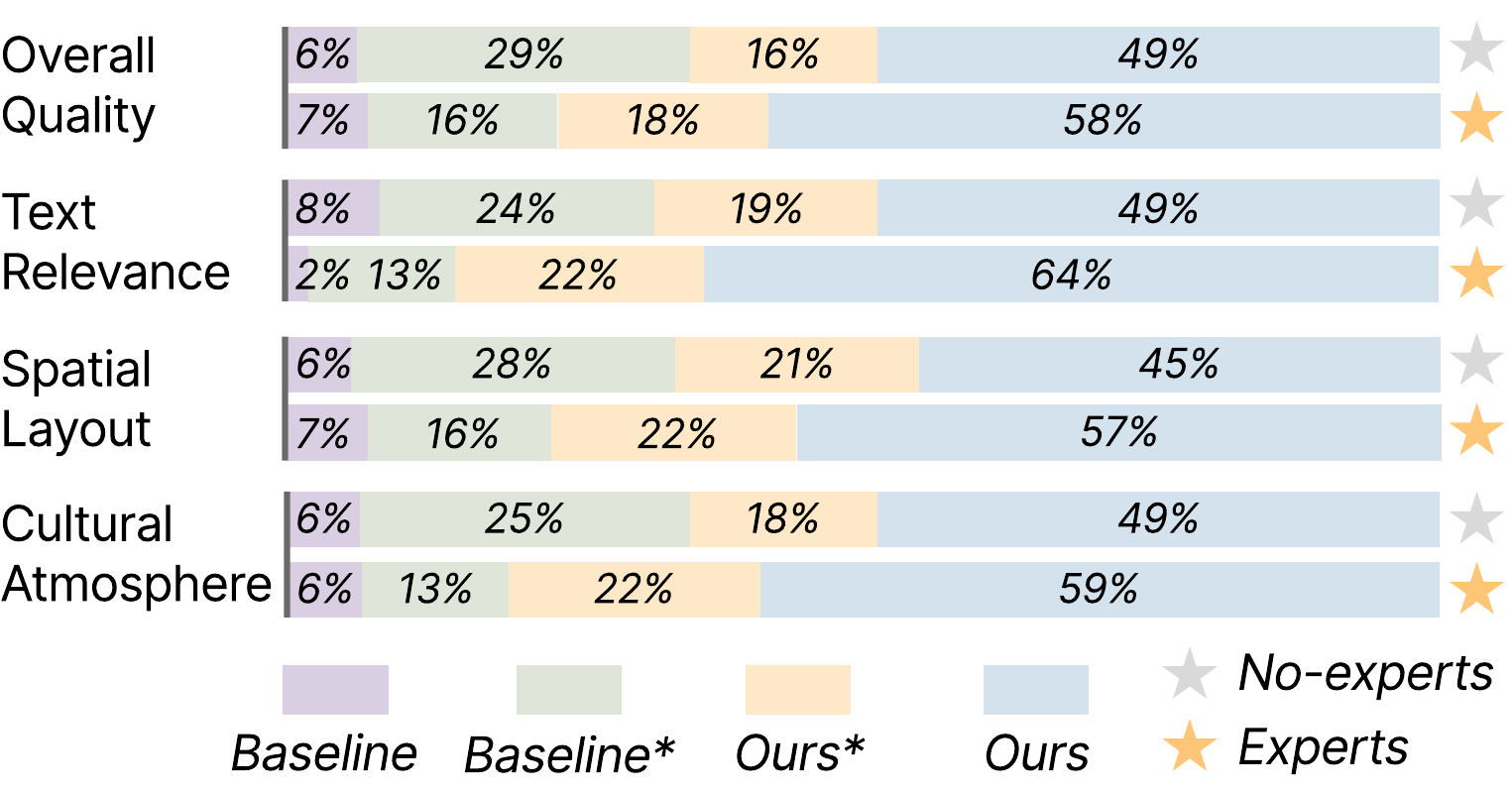}
  \caption{Comparing the selection ratio of four methods in the experiment from four perspectives: (1) Overall Quality, (2) Text Relevance, (3) Spatial Layout, (4) Cultural Atmosphere.}
  \label{fig: supp user study}
\end{figure}

\subsection{Human Evaluation}
We invited 11 garden experts and 32 non-expert volunteers to evaluate the aesthetic quality of the generated Jiangnan gardens. Jiangnan gardens for human evaluation comprise five types
in the Table~\ref{tab:supp ablation in user study}. 
With the chain of agents and knowledge integration, \textbf{humans prefer GardenDesigner over baseline methods} from all perspectives.
The baseline method receives fewer selections (all under 10\%) compared to other methods adopted with GardenVerse. On the contrary, the baseline method gets more preference using the GardenVerse datasets, 
indicating that \textbf{GardenVerse promotes the whole garden quality}. 
We also removed the Knowledge-Embedded Asset Arrangement module to conduct ablation study. Although two scenes have the same terrain and structure layout, \textbf{layout with aesthetic rules gets more preference}, indicating that aesthetic principles play a significant role in determining scene quality.

\subsection{Discussion}
The chain of agents
has the potential to generalize to other artistic scene generation tasks.
First, by encoding new scene rules and knowledge in textual form, the knowledge-Embedded context mechanism can be directly reused by vectorizing them into semantic memory space.
Second, terrain and path generation agents can be adapted to various landscape typologies by modifying the procedural loss terms and path-scoring rules. For example, European royal gardens can be generated by imposing symmetry-aware optimization loss and balanced path scoring.

\section{Applications}
\begin{figure}[th]
  \centering
  \includegraphics[width=0.475\textwidth]{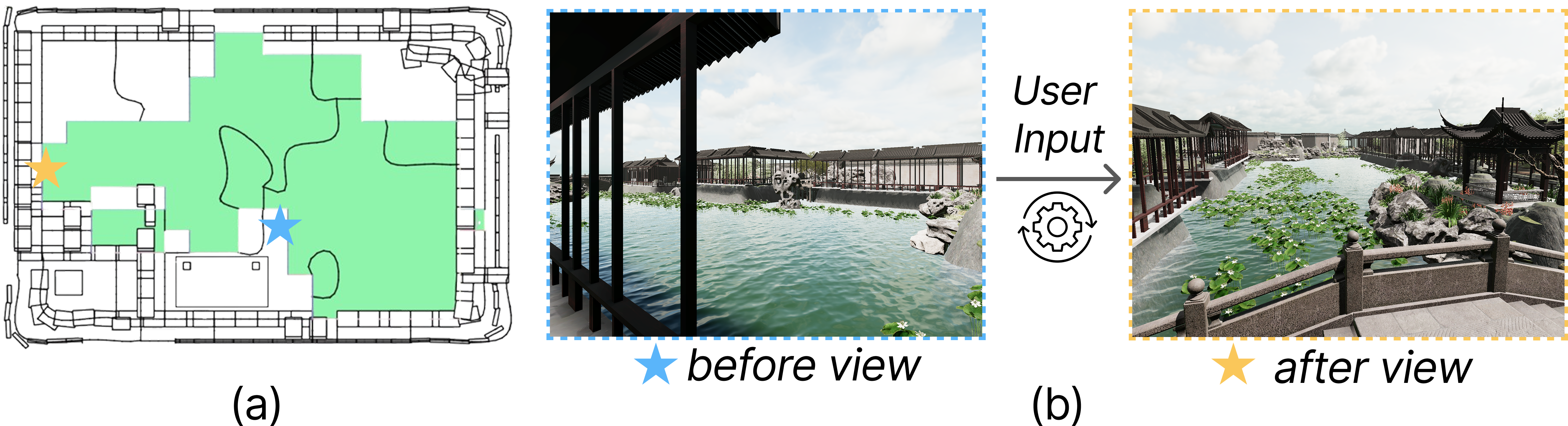}
  \caption{Two applications: (a) Generating a 2D garden construction layout, which can be used to build the garden; (b) Navigating to a spot of interest following the user's instructions. 
  }
  \label{fig: application}
\end{figure}

We developed an interface to allow users to input text and construct a Jiangnan garden in Unity.
We also provide a terrain adjustment tool to modify the terrain and output the structure map to assist engineers in the construction of physical gardens. Furthermore, users can input instructions to navigate to a spot of interest in the garden using VLMs, as shown in Figure~\ref{fig: application}. 
Our GardenDesigner system can support Jiangnan garden design, virtual tourism, interactive entertainment, and virtual reality experiences.
  
 
%% file: Sec/6_conclusion.tex
\section{Conclusion} 
This paper has proposed \textbf{GardenDesigner}, a novel framework that encodes aesthetic principles for Jiangnan garden construction and integrates a chain of agents procedurally.
By structuring the generation process into hierarchical garden composition and knowledge-embedded asset arrangement, GardenDesigner ensures spatial rationality and aesthetic coherence with Jiangnan garden design principles.
Looking forward, GardenDesigner can be extended to support interactive educational tools, virtual heritage reconstruction, and personalized landscape design, opening new avenues for cultural heritage preservation and creative applications in digital art and games.

%% file: Sec/7_supp.tex
\maketitlesupplementary

\section{Implementation Details}
\subsection{Terrain Generation Algorithm}
To generate the garden's terrain, we adopt the genetic algorithm on a 2D grid. We initialize the terrain with a random integer from 0 to 3, representing unused area, water area, land area, and ground area.
Roads are selected and combined from the grid borders, which are scored through border positions. 
Roads will be smoothed through spline curving and choosing the intersection or inflection point as the start and end of the curve.
The CLIP model we used to evaluate CLIP-Score is openai/clip-vit-base-patch32~\cite{clip}. 

\subsection{Hierachical Garden Composition}
\begin{figure*}[ht]
  \centering
  \includegraphics[width=0.9\textwidth]{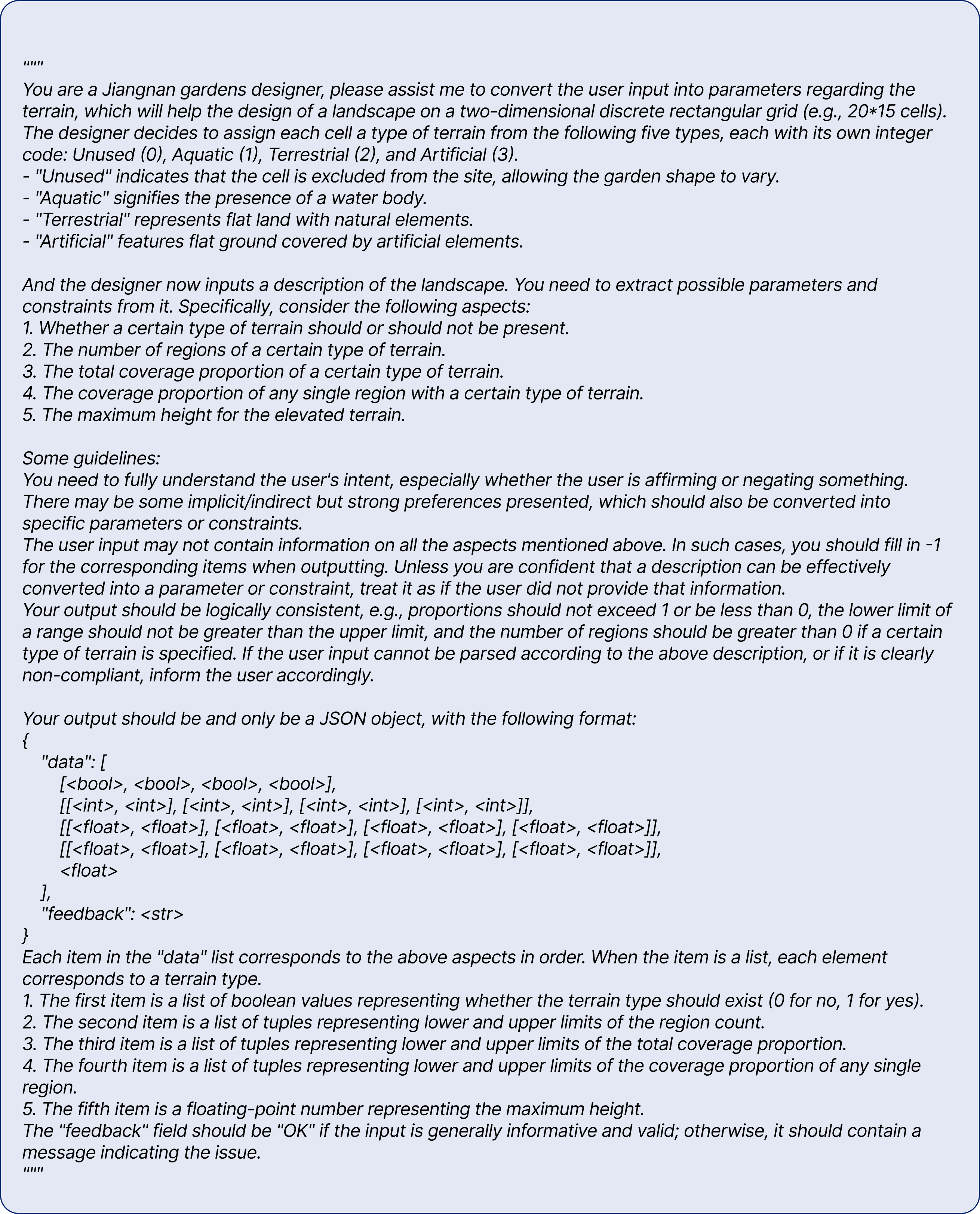}
  \caption{Prompts for the terrain generation agent.}
  \label{fig:supp terrain generation}
\end{figure*}

\begin{figure*}[ht]
  \centering
  \includegraphics[width=0.9\textwidth]{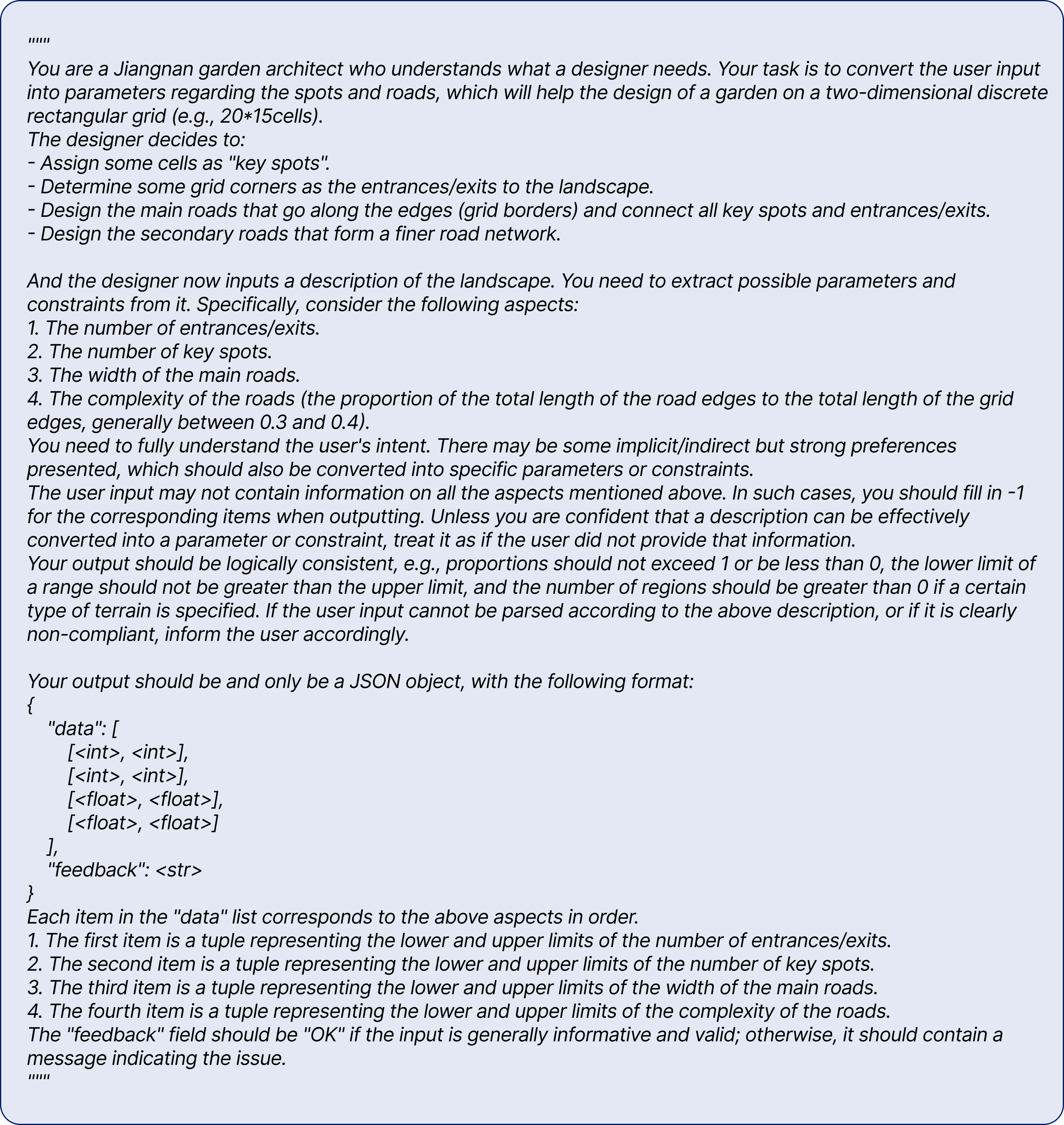}
  \caption{Prompts for the road generation agent.}
  \label{fig:supp structure generation}
\end{figure*}

In the implementation process, we divide the Chain of Aesthetic Principles into terrain and structure generation, using a genetic algorithm. Firstly, the detailed prompts of the \textbf{Terrain Distribution Agent ($\mathcal{A}_T$)} are presented in Figure~\ref{fig:supp terrain generation}. Additionally, we also provide the detailed prompts of the \textbf{Road Generation Agent ($\mathcal{A}_R$)} in Figure~\ref{fig:supp structure generation}. Based on the response from LLM, we parse the parameters for 2D genetic alogrithm to generate terrain and structures. 
We choose four types of terrains to simulate the landform of Jiangnan garden: \textit{Outside}, \textit{Waterbody}, and \textit{Land}. Specifically, each terrain is explained as follows:
\noindent\begin{itemize}
  \item \textbf{Outside} areas refer to unoccupied zones and serve to increase spatial diversity and boundary complexity.
  \item \textbf{Waterbody} areas present the indispensable and symbolic water area of the Jiangnan garden.
  \item \textbf{Land} areas represent flat land with natural elements.
  \item \textbf{Ground} areas are the flat terrain zone, on which the buildings, plants, and rocks will be sited.
\end{itemize}
In each terrain grid cell, we employ an integer number (0-3) to represent these terrain types.

\begin{algorithm}[t]
\caption{Garden Construction}
\label{alg:garden_synthesis}
\begin{algorithmic}[1]
\REQUIRE User input text $U$, Garden Principles $K_\text{global}$, Asset library $O_\text{asset}$ with knowledge $K_\text{a}$
\ENSURE Complete garden $G = (T, R, (O_\text{s}, P)$

\STATE $T \gets A_T(U, K_\text{global})$ 
\STATE $R \gets A_\text{R}(S(T, e_{i,j}), U, K_\text{global})$
\STATE $O_\text{s} \gets A_\text{S}(Q((V(K_\text{a}), o_i), U), I_\text{area})$
\STATE $C \gets A_C(Q((V(K_\text{a}), o_i, o_j), U))$

\FOR{each position $p_\text{anchor}$ for $o_\text{anchor}$}
    \STATE $P_\text{temp} \gets \{p_\text{anchor}\}$
    \IF{$\text{DFS-Place}(O_s,o_\text{anchor}, C, P_\text{temp})$}
        \IF{$L_\text{opt}(P_\text{temp}) < L_\text{opt}(P)$}
            \STATE $P \gets P_\text{temp}$
        \ENDIF
    \ENDIF
\ENDFOR
\RETURN $G = (T, R, (O_\text{s}, P))$

\end{algorithmic}
\end{algorithm}

\subsection{Knowledge-embedded Asset Arrangement}
\begin{figure*}[ht]
  \centering
  \includegraphics[width=0.9\textwidth]{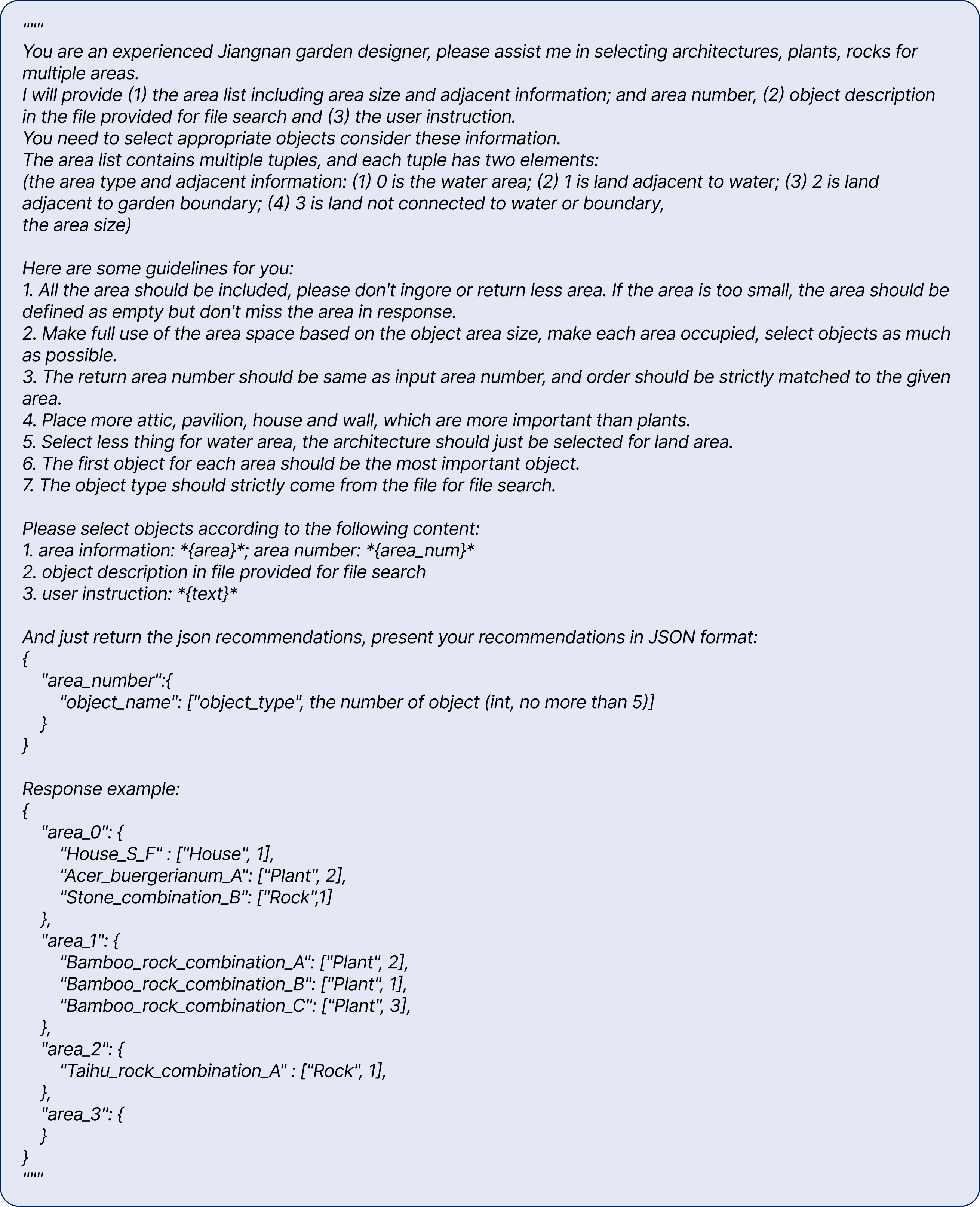}
  \caption{Prompts for the asset selection agent.}
  \label{fig:supp landscape planner}
\end{figure*}

\begin{figure*}[ht]
  \centering
  \includegraphics[width=0.9\textwidth]{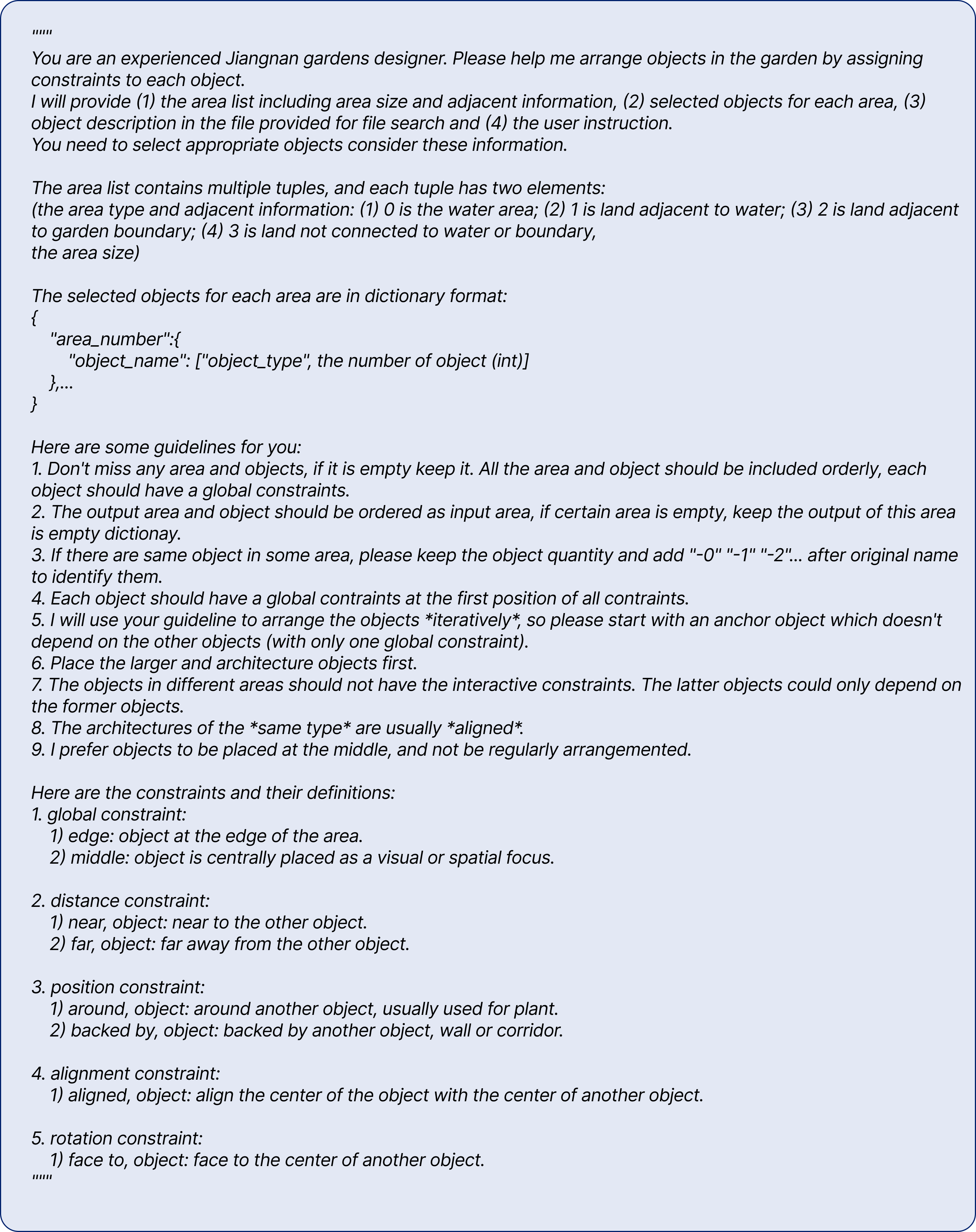}
  \caption{First part of prompts for the layout
optimization agent.}
  \label{fig:supp constraint_1}
\end{figure*}

\begin{figure*}[ht]
  \centering
  \includegraphics[width=0.9\textwidth]{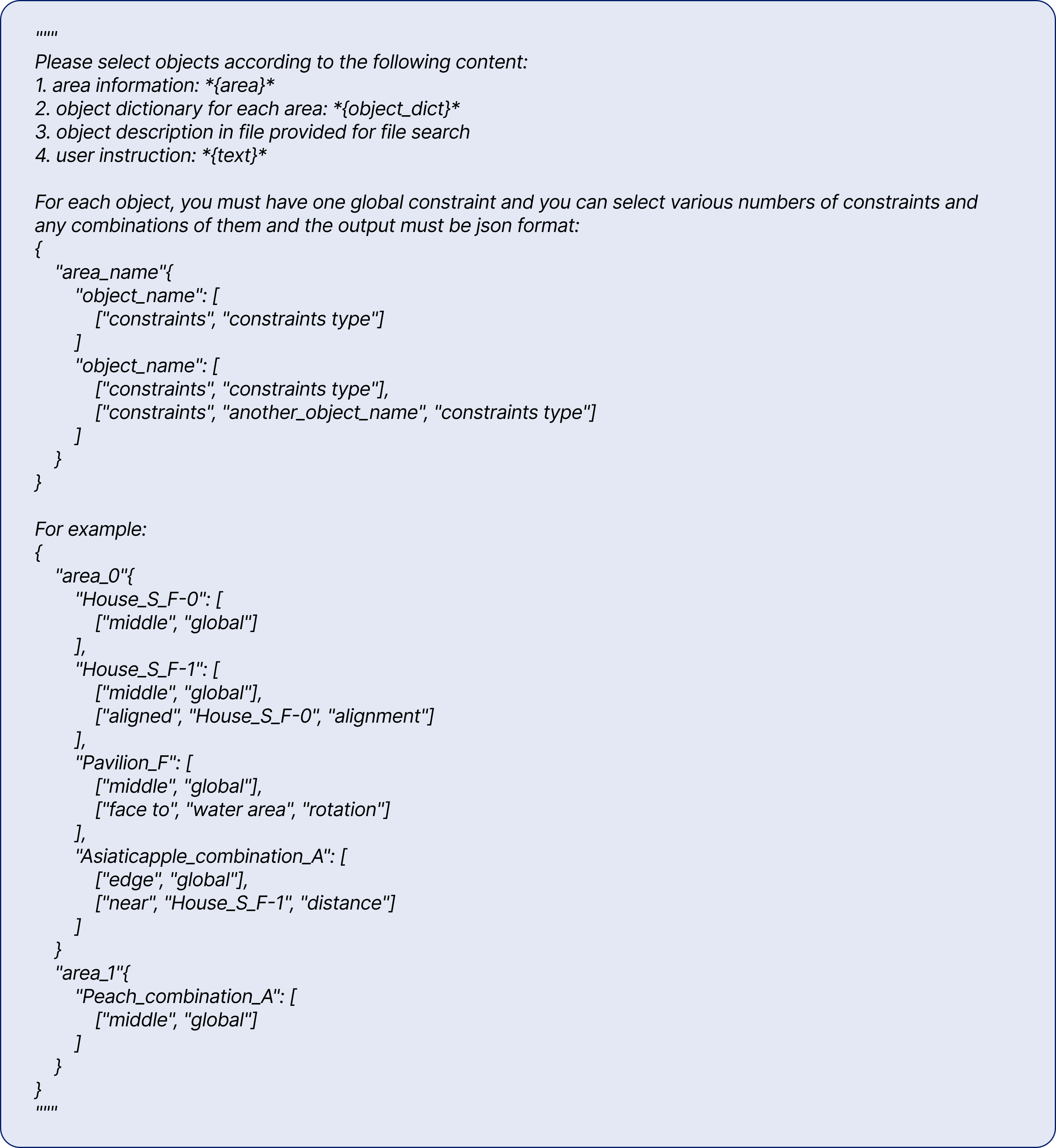}
  \caption{Second part of prompts for the layout
optimization agent.}
  \label{fig:supp constraint_2}
\end{figure*}

To obtain an appropriate garden layout, we decompose the Garden Configuration into object selection and constraint setting. We present the detailed prompt used to select objects for \textbf{Asset Selection Agent ($\mathcal{A}_S$)} in Figure~\ref{fig:supp landscape planner}. Before requesting LLM, we annotate each area with area information. We use the file search tools from OpenAI. After selecting the appropriate objects, we also present the \textbf{Layout Optimization Agent ($\mathcal{A}_C$)} prompt in Figure~\ref{fig:supp constraint_1} and Figure~\ref{fig:supp constraint_2} , enabling the feasible constraints for objects. All constraints are formalized in a structured representation to ensure interpretability and implementation feasibility: 
{\small
\begin{verbatim}
"area name": {
    "object name": [
        ["constraint", "type"],
        ["constraint", "rel object", "type"]
    ]
}
\end{verbatim}
}
\normalsize
where the ``area name'' and ``object name'' are the target area and object, the ``constraint'' is the relationship between object and another object with the name of ``rel object'', and the ``type'' is the constraint type.

\subsection{Optimization}
We demonstrate the detailed information in Optimization section. We utilize Depth-First Search (DFS) Solver to optimize object constraints from Garden Configuration inspired by~\citet{holodeck}. 
To make the balance between time and quality, we choose to change the area into grid point according to the area bounding box and we also remove the points of the area. The grid points in the area are presented as the solution for each object position movement. 
In the DFS solver, each object is characterized by five variables: $(x, y, l, w, rotation)$, where: $(x, y)$ represents the 2D coordinates of the object’s center, $l$ and $w$ denote the length and width of the object’s 2D bounding box. Rotation can take one of four possible angles: 0, 90, 180, or 270, where 0 is forward positive z-direction. The solver applies soft constraints, permitting minor violations to facilitate feasible layout generation. Apart from object constraint, we also hard constraints are enforced to ensure physically valid placements: (1) No object collisions, objects must not overlap; (2) Area boundaries, objects must stay within the designated space. If an object violates any hard constraint, it is rejected from the current layout. We calculate the overall loss of each objects in validate solution, and select the most feasible solution with lowest loss after 100 iteration steps. 

\section{GardenVerse Details}
\begin{figure}[th]
  \centering
  \includegraphics[width=0.5\textwidth]{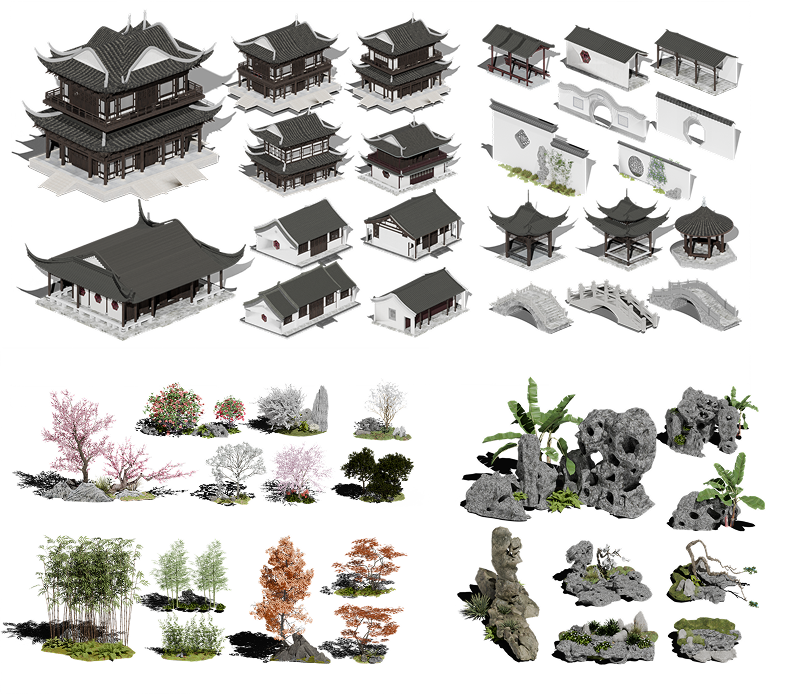}
  \caption{GardenVerse data examples. The GardenVerse consists of four types of objects: (a) the Architecture, (b) the Structure, (c) the Plant, and (d) the Rock. The plant and rock include both single and combined objects.}
  \label{fig:supp GardenVerse}
\end{figure}

\begin{figure}[th]
  \centering
  \includegraphics[width=0.40\textwidth]{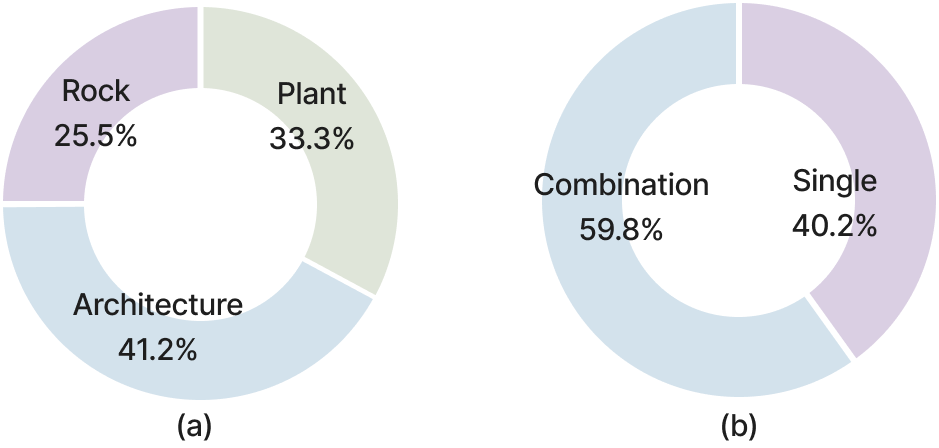}
  \caption{GardenVerse statistics: (a) the object categories proportional distribution; (b) the combined and single objects ratio.}
  \label{fig:supp data_statistics}
\end{figure}

GardenVerse comprises 132 high-quality artistic 3D assets across three canonical categories: Rock (33), Plant (44), and Architecture (54), in Figure~\ref{fig:supp GardenVerse}. In Jiangnan gardens, the combination of plants and rocks stands out as a distinctive feature compared to standalone assets, which creates a harmonious interplay between organic vitality and enduring solidity. It includes both individual elements ($40.2\%$) and pre-composed arrangements ($59.8\%$) of plants and rocks, enabling flexible retrieval of Jiangnan gardens, in Figure~\ref{fig:supp data_statistics}.

\small
\begin{verbatim}
{
    "name": "object name",
    "path": "related path",
    "pos": "appropriate position",
    "object": "internal object",
    "season": "appropriate season",
    "description": "knowledge about object",
    "minp": "min position",
    "maxp": "max position",
    "size": "object size"
}
\end{verbatim}
\normalsize
where the text of ``description'', ``season'' and ``pos'' constitute the asset garden knowledge to guide the asset selection and garden layout optimization.

\begin{figure*}[th]
  \centering
  \includegraphics[width=0.935\textwidth]{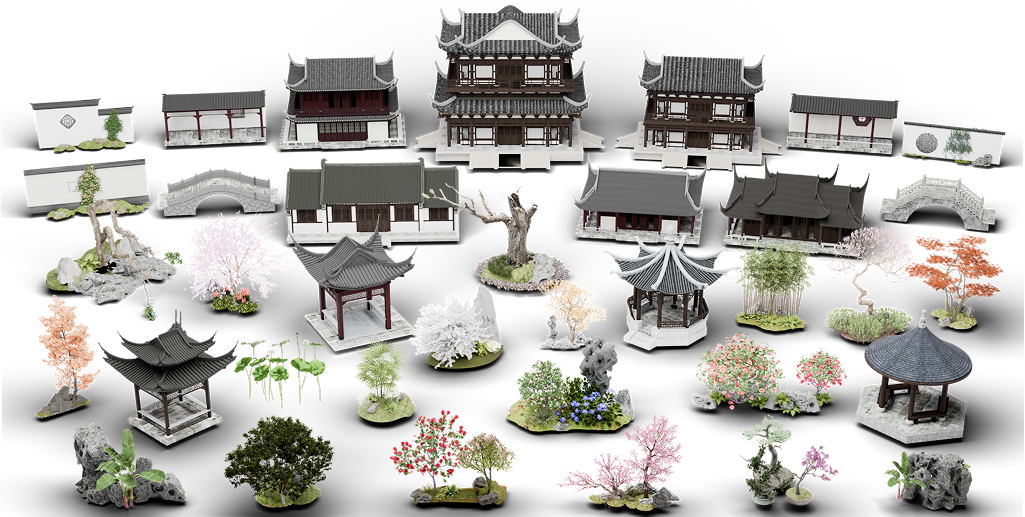}
  \caption{GardenVerse dataset. GardenVerse includes a collection of diverse 3D objects specially designed for Jiangnan gardens, and we show object examples from the dataset. GardenVerse encompasses four distinct object categories: architecture, plant, and rock, containing single objects and combination asset forms.}
  \label{fig:data_statistics}
\end{figure*}

\section{Experiments}
\begin{figure*}[ht]
  \centering
    \includegraphics[width=0.93\textwidth]{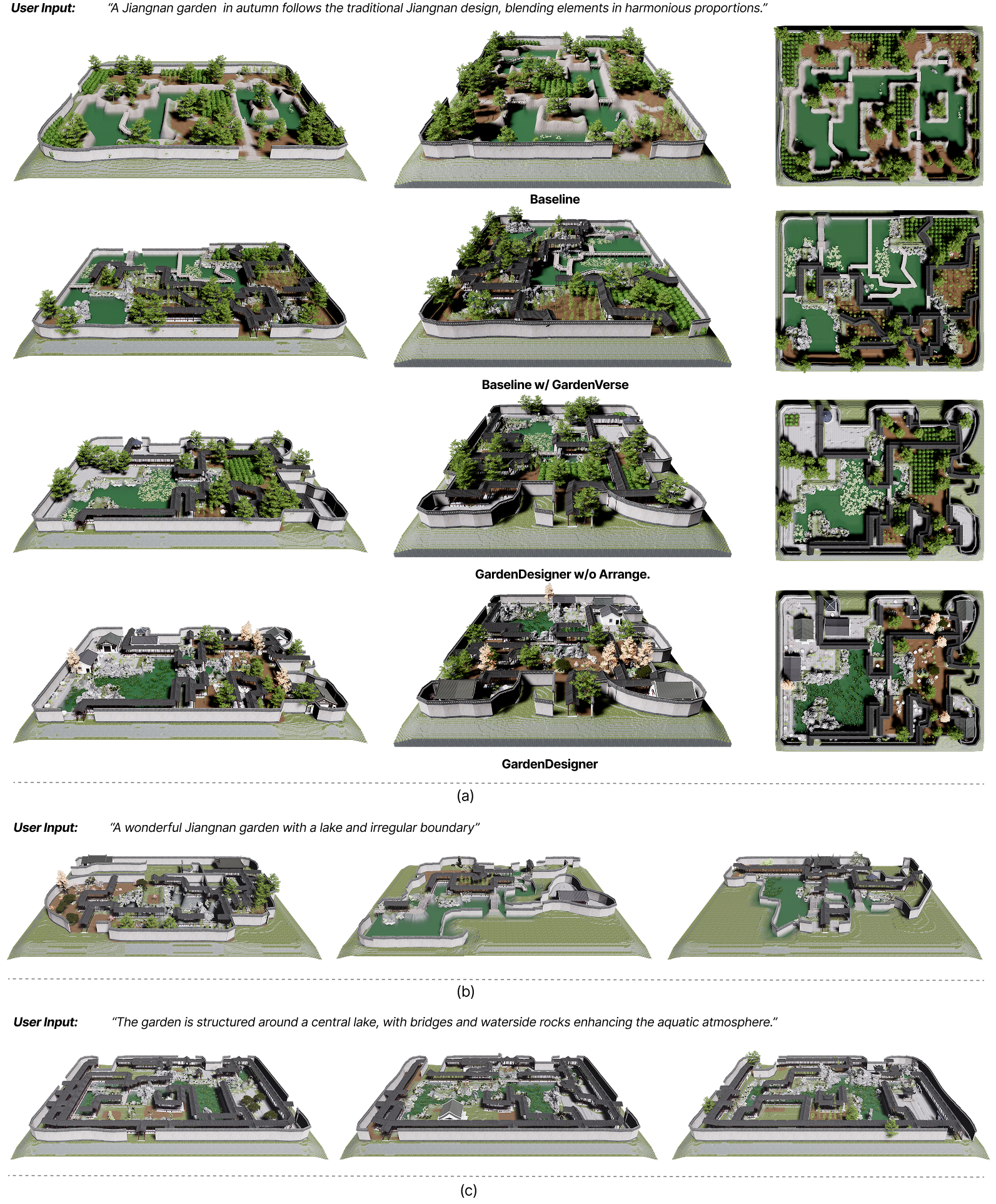}
  \caption{Qualitative results. (a) We conduct the ablation study to compare the different methods for garden construction with same user input, and the view from left to right is front view, right view and top view. (b) We input the same user instruction to evaluate the generation diversity of GardenDesigner. (c) We also input the same user instruction and keep the same terrain to generate different gardens.}
  \label{fig:supp qualitative comparison}
\end{figure*}

We conducted the ablation study in Figure~\ref{fig:supp qualitative comparison}(a). The baseline method has worse visual quality than other methods with GardenVerse. The methods with terrain loss and  explorative road scoring function have water-centric terrain and reasonable pathway.
Furthermore, we evaluate the diversity of GardenDesigner. In Figure~\ref{fig:supp qualitative comparison}(b), we input the same prompt and evaluate
the diversity of GardenDesigner to generate different Jiangnan gar-
dens. In Figure~\ref{fig:supp qualitative comparison}(c), we evaluate the object layout diversity through
maintaining the same prompt, terrain, and structure layout.

Table~\ref{tab: rebuttal quantitative comparison} shows that the comparison with natural scene generator Infinigen~\cite{infinigen}, rule-based SceneX~\cite{zhou2024scenex}, diffusion-based NuiScene~\cite{NuiScene}, and PCG method without chain of agents. GardenDesigner outperforms others on all consistency and aesthetic metrics, validating the effectiveness. Based on the performance of baseline and PCG methods, both LLM-based reasoning and procedural engineering contribute to the improvement.
\begin{table}[th]
  \caption{Quantitative comparison with different methods.}
  \centering
  \small
  \begin{tabular*}{0.475\textwidth}{@{\extracolsep{\fill}}llllll@{}}
    \toprule
     Method 
     & \multicolumn{1}{c}{CLIP-S $\uparrow$} 
     & \multicolumn{1}{c}{CLIP-A $\uparrow$} 
     & \multicolumn{1}{c}{VLM-S $\uparrow$} 
     & \multicolumn{1}{c}{QA-Quality$ \uparrow$} 
     \\  \midrule 
    Infinigen
    & \multicolumn{1}{c}{18.1}
    & \multicolumn{1}{c}{51.6} 
    & \multicolumn{1}{c}{6.3} 
    & \multicolumn{1}{c}{24.9}\\
    SceneX
    & \multicolumn{1}{c}{23.1}
    & \multicolumn{1}{c}{53.1} 
    & \multicolumn{1}{c}{5.5} 
    & \multicolumn{1}{c}{37.6}\\
    NuiScene
    & \multicolumn{1}{c}{25.6}
    & \multicolumn{1}{c}{53.7} 
    & \multicolumn{1}{c}{10.3} 
    & \multicolumn{1}{c}{46.3}\\
    PCG
    & \multicolumn{1}{c}{27.4}
    & \multicolumn{1}{c}{53.9} 
    & \multicolumn{1}{c}{28.9} 
    & \multicolumn{1}{c}{51.2}\\
    Ours 
    & \multicolumn{1}{c}{\textbf{27.6}} 
    & \multicolumn{1}{c}{\textbf{54.2}} 
    & \multicolumn{1}{c}{\textbf{32.5}} 
    & \multicolumn{1}{c}{\textbf{53.8}}\\
  \bottomrule
  \end{tabular*}
  \label{tab: rebuttal quantitative comparison}
\end{table}

We conduct additional loss ablation study to better understand the contribution of different losses. Table~\ref{tab: rebuttal ablation} shows that all losses will affect the garden structure complexity and visual quality, while global and distance losses contribute significantly to visual quality in VLM-S and QA-Quality. The weights for loss components are defined by their contribution to the final performance.

\begin{table}[th]
  \caption{Ablation study results on optimization losses.}
  \centering
  \small
  \begin{tabular*}{0.4\textwidth}{@{\extracolsep{\fill}}llllll@{}}
    \toprule
     Method & \multicolumn{1}{c}{ FD } & \multicolumn{1}{c}{VLM-S $\uparrow$} & \multicolumn{1}{c}{QA-Quality $\uparrow$}  
     \\  \midrule 
    w/o $L_{glo}$
    & \multicolumn{1}{c}{1.39} 
    & \multicolumn{1}{c}{31.8}  
    & \multicolumn{1}{c}{48.9}\\
    w/o $L_{pos}$
    & \multicolumn{1}{c}{1.38} 
    & \multicolumn{1}{c}{32.2} 
    & \multicolumn{1}{c}{50.9}\\
    w/o $L_{dis}$
    & \multicolumn{1}{c}{1.38} 
    & \multicolumn{1}{c}{31.9}
    & \multicolumn{1}{c}{48.6}\\
    w/o $L_{ali}$
    & \multicolumn{1}{c}{1.40} 
    & \multicolumn{1}{c}{32.3} 
    & \multicolumn{1}{c}{49.6}\\
    w/o $L_{rot}$
    & \multicolumn{1}{c}{1.36}
    & \multicolumn{1}{c}{32.3}
    & \multicolumn{1}{c}{50.3}\\
    Ours 
    & \multicolumn{1}{c}{\textbf{1.36}}
    & \multicolumn{1}{c}{\textbf{32.5}} 
    & \multicolumn{1}{c}{\textbf{53.8}} \\
  \bottomrule
  \end{tabular*}
  \label{tab: rebuttal ablation}
\end{table}

\section{Human Evaluation Details}

\subsection{Human Evaluation Setup}
\begin{figure*}[ht]
  \centering
    \includegraphics[width=0.9\textwidth]{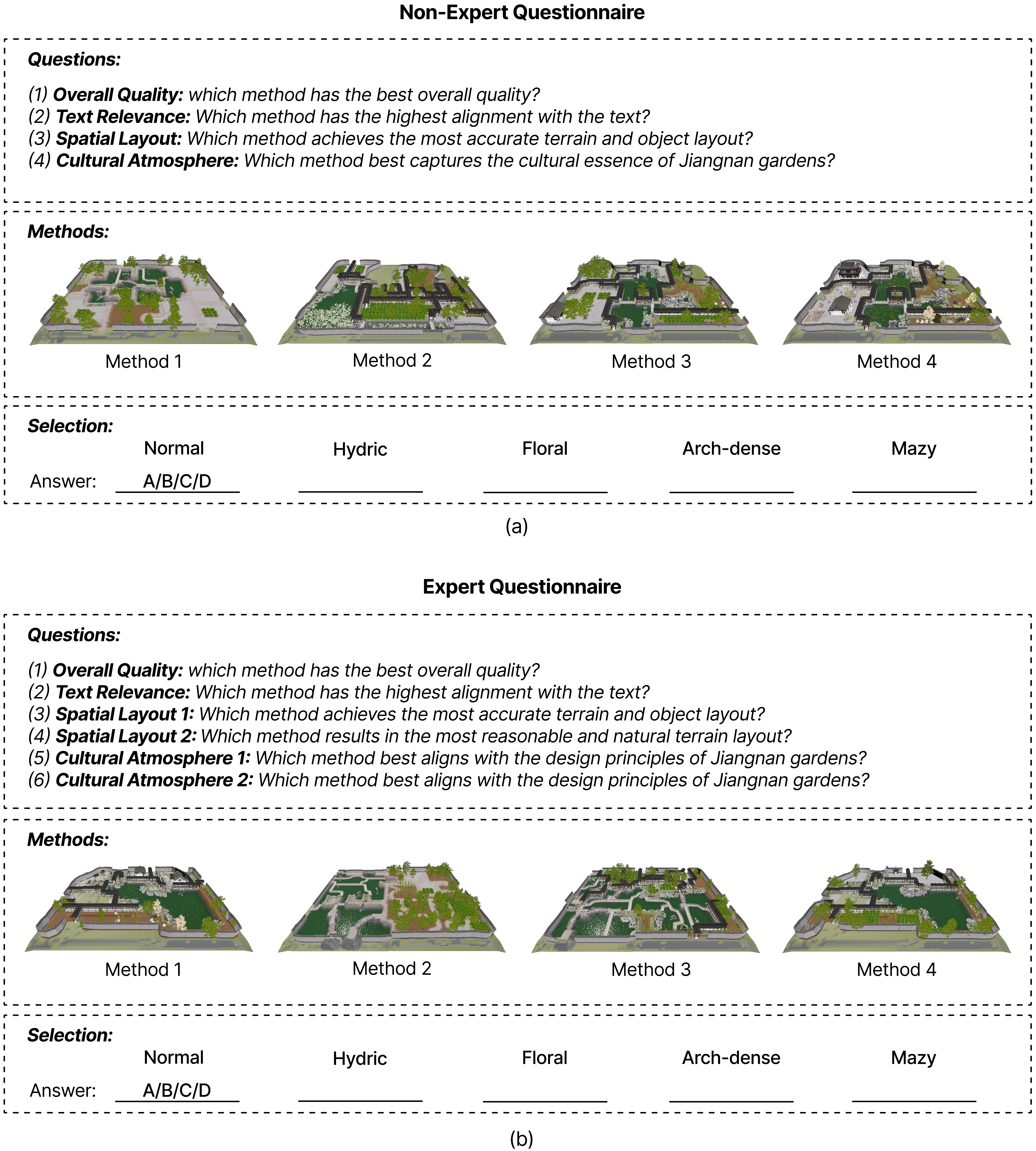}
  \caption{Questionnaire Survey. We conducted human evaluation with experts and no-experts. (a) In non-expert questionnaire, we provide four questions from overall quality, text relevance, spatial layout, and cultural atmosphere for volunteers to answer. And we provide five types of generated Jiangan gardens from four different methods. (b) We refine the question about sptatial layout and cultural atmosphere.}
  \label{fig:supp question}
\end{figure*}

We also invite 11 garden experts and 32 non-expert volunteers to evaluate the aesthetic quality of the generated Jiangnan gardens in Figure~\ref{fig:supp question}. We prepared 20 Jiangnan gardens for human evaluation, comprising five types of garden: (1) Normal, (2) Hydric, (3) Floral, (4)Arch-dense , and (5) Mazy. We ask the volunteers to choose which Jiangnan garden is better based on four perspectives: (1) \textbf{Overall Quality}: which method has the best overall quality? (2) \textbf{Text Relevance}: Which method has the highest alignment with the text? (3) \textbf{Spatial Layout}: Which method achieves the most accurate terrain and object layout? (4) \textbf{Cultural Atmosphere}: Which method best captures the cultural essence of Jiangnan gardens? For experts, we add more detailed questions. For Spatial Layout, we add two questions: (1) Which method results in the most reasonable and natural terrain layout? (2) Which approach provides the most logical and organic arrangement for vegetation and structures?. For Cultural Atmosphere, we also add two questions: (1) Which method best aligns with the design principles of Jiangnan gardens? (2) Which method best captures the poetic essence and philosophical depth of Jiangnan gardens?
\subsection{Human Evaluation Results}
\textbf{Humans prefer GardenDesigner over baseline}. Humans prefer the gardens generated from GardenDesigner compared to other methods, with a majority of selection, especially Cultural Atmosphere (49\% General Users, 59\% Experts).
Overall Quality (49\% General Users, 58\% Experts), Text Relevance (49\% General Users, 64\% Experts), Spatial Layout (45\% General Users, 57\% Experts) and Cultural Atmosphere (49\% General Users, 59\% Experts). 

\textbf{GardenVerse promote the whole garden quality}. The baseline method receives few selection compared to other methods adopted with GardenVerse. The selection ratios for all the question in General Users and Experts are all under 10\%. On the contrary, the baseline method gets more preference using the GardenVerse datasets, especially among General Users. It gets 22\% more selection ratio than the baseline method~\cite{conlan}, validating the effects of GardenVerse. 

\textbf{Layout with aesthetic rules gets more preference}. We also conducted an ablation study about Garden Configuration. We modify GardenDesigner by removing the Garden Configuration module. Although two scenes have the same terrain and structure layout, the Humans prefer GardenDesigner more, indicating that Garden Configuration plays a significant role in determining scene quality.

\section{Applications}
To visualize the garden, we provide the plugin in Unity, where the user can edit and interact in real time.
The output of GardenDesigner are stored as files containing all necessary information: the height map for terrain generation, textures to distinguish different terrains, and object information in json format, in which all of these constructs the Jiangnan Garden. We also develop Unity plug-in is developed to parse these files and convert them into terrain and objects. 
Additionally, we also provide the terrain adjustment tool to modify the terrain boundary and output the structure map to assist garden design and building.



\clearpage

\clearpage

\clearpage
\clearpage
\clearpage
\clearpage
\clearpage
